%% file: arxiv.tex
\documentclass[10pt,twocolumn,letterpaper]{article}

\PassOptionsToPackage{table}{xcolor}
\usepackage[pagenumbers]{cvpr}

\input{preamble}

\usepackage{float}

\definecolor{cvprblue}{rgb}{0.21,0.49,0.74}
\usepackage[pagebackref,breaklinks,colorlinks,allcolors=cvprblue]{hyperref}

\def\confName{CVPR}
\def\confYear{2027}

\title{Trainspotting: Learning Object-Centric 3D Reconstruction from Found Photos}

\author{Hanyu Chen \qquad Noah Snavely\\
Cornell University}

\begin{document}

\twocolumn[{%
\renewcommand\twocolumn[1][]{#1}%
\maketitle
\input{sec/0_teaser}
}]

\input{sec/0_abstract}
\input{sec/1_intro}
\input{sec/2_related}
\input{sec/3_dataset}

\input{sec/4_pseudogt}
\input{sec/6_experiments}

\input{sec/8_conclusion}
\input{sec/9_acknowledgments}

{
    \small
    \bibliographystyle{ieeenat_fullname}
    \bibliography{references}
}

\clearpage
\appendix
\renewcommand{\thesection}{\Alph{section}}
\renewcommand{\thetable}{S\arabic{table}}
\renewcommand{\thefigure}{S\arabic{figure}}
\setcounter{section}{0}\setcounter{table}{0}\setcounter{figure}{0}
\renewcommand{\floatpagefraction}{0.5}
\renewcommand{\dblfloatpagefraction}{0.5}
\setcounter{dbltopnumber}{9}
\twocolumn[{\centering\Large\bfseries Supplementary Material\par\vspace{1.2em}}]

\input{sec/S1_curation}

\input{sec/S2_pipeline}
\input{sec/S3_training}
\input{sec/S4_evaluation}
\input{sec/S6_qualitative}
\input{sec/S5_results}

\input{sec/S7_prompts}
\input{sec/S8_floats}

\end{document}

%% file: preamble.tex
\input{math_commands}

\usepackage{tabularx}
\usepackage{tikz}
\usepackage{ragged2e}
\usetikzlibrary{arrows.meta}
\usepackage[subtle]{savetrees}
\makeatletter
\def\cvprsection{\@startsection{section}{1}{\z@}{-8pt plus -2pt minus -2pt}{7pt}{\large\bf}}
\def\cvprsubsection{\@startsection{subsection}{2}{\z@}{-6pt plus -1pt minus -1pt}{5pt}{\elvbf}}
\def\cvprsubsubsection{\@startsection{subsubsection}{3}{\z@}{-6pt plus -2pt minus -2pt}{3pt}{\tenbf}}
\makeatother

\newcommand{\datasetname}{\textsc{Trainspotting}}
\newcommand{\modelname}{\textsc{Spotter}}

%% file: math_commands.tex
\usepackage{amsmath,amsfonts,bm}

\def\eqref#1{equation~\ref{#1}}

\def\1{\bm{1}}

\def\vg{{\bm{g}}}

\def\vp{{\bm{p}}}

\def\vx{{\bm{x}}}

\def\mF{{\bm{F}}}

\DeclareMathAlphabet{\mathsfit}{\encodingdefault}{\sfdefault}{m}{sl}
\SetMathAlphabet{\mathsfit}{bold}{\encodingdefault}{\sfdefault}{bx}{n}



%% file: sec/0_teaser.tex
\begin{center}
\captionsetup{type=figure}
\vspace{-1em}
\includegraphics[width=\textwidth]{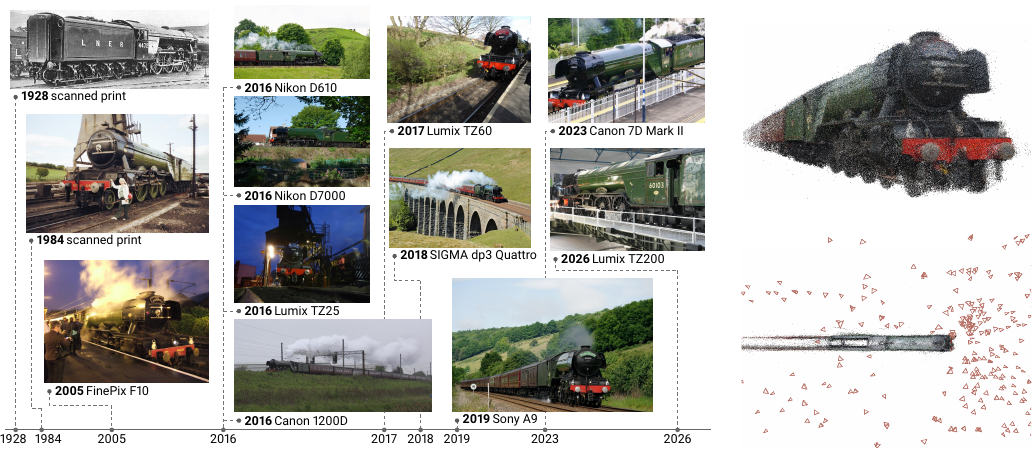}
\caption{\textbf{One object over one century.}
Wikimedia Commons holds 727 photos of the locomotive \emph{Flying Scotsman}, taken from the 1920s to 2026 by 187 photographers using 123 camera models.
The twelve photos on the left are labeled with their year and camera; our recovered sparse point cloud and camera poses are shown on the right.
\datasetname{} contains 12{,}068 such objects reconstructed from 311{,}886 photos from Wikimedia Commons.
We use this dataset to substantially improve feedforward 3D reconstruction models on in-the-wild object-centric photo collections.}
\label{fig:teaser}
\end{center}\vspace{0.5em}

%% file: sec/0_abstract.tex
\begin{abstract}
State-of-the-art feedforward reconstruction models perform poorly on in-the-wild object-centric photo collections.
They are trained largely on single-session object captures whose images vary in viewpoint while the object and its environment remain nearly fixed.
Public photo archives contain a different form of multi-view data.
They hold \emph{found photos} of the same object or design: photos taken independently by many people across years, environments, cameras, and sometimes physical units.
In this work, we introduce {\upshape\datasetname{}}, a dataset of 12{,}068 objects and 311{,}886 photos from Wikimedia Commons.
We reconstruct each object to recover camera poses, multi-view stereo depth, and object masks.
We categorize the objects into three regimes: objects fixed at one site, objects that move between sites, and designs shared by many physical units.
Evaluating recent feedforward reconstruction models on {\upshape\datasetname{}} shows that they fail most severely in the last two regimes.
Fine-tuning VGGT-$\Omega$ on {\upshape\datasetname{}} significantly improves camera pose, depth, and 3D geometry on held-out objects.
We observe similar improvements on the independent NAVI in-the-wild benchmark, which contains photos of each object in different environments.
Performance on conventional benchmarks remains largely unchanged.
\end{abstract}

%% file: sec/1_intro.tex
\section{Introduction}
\label{sec:intro}

\emph{
Choose life.
Choose a train, a statue, a pipe organ.
Choose a 1967 Volkswagen Beetle, a K6 red telephone box in London.
We chose not to choose.
We reconstructed them all.}

Found photos of objects like those listed above accumulate in public archives such as Wikimedia Commons over many years.
Feedforward reconstruction models~\citep{wang2025vggt,wang2026pi3,wang2026vggtomega}, however, obtain most of their object-level training data from datasets that record each object in a single video or capture session~\citep{reizenstein2021co3d,xia2024wildrgbd,liu2025uco3d}.
These captures densely sample nearby views while the object and its environment remain nearly fixed.

\begin{figure}[t]
\centering
\includegraphics[width=\columnwidth]{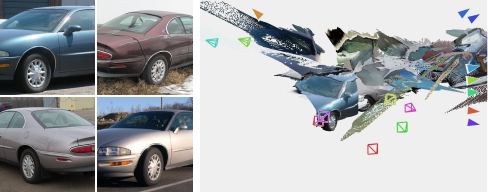}
\caption{\textbf{VGGT-$\Omega$ fails on the \emph{Buick Riviera}.}
We visualize the camera poses and point map predicted by VGGT-$\Omega$ from sixteen photos of the Buick Riviera (multiple physical instances of the same car design).
The views are misregistered and the car is reconstructed as scattered fragments.
Figure~\ref{fig:qualitative} shows the reconstruction of our fine-tuned model from the same photos.}
\label{fig:failure}
\vspace{-1em}
\end{figure}

The photos of an object on Wikimedia Commons follow no shared capture plan.
Their viewpoints are unevenly sampled, and their environment may change dramatically from one photo to the next (e.g., a ship on the open seas vs.\ the same ship in port).
An individual object may be repainted or rebuilt over time, and photos of a mass-produced product, like a specific make and model of car, may depict different physical instances.
Datasets of \emph{landmarks} and other places have been built from similar in-the-wild Internet photo collections~\citep{snavely2006phototourism,li2018megadepth,tung2024megascenes}.
However, these scene-level photo collections depict a largely consistent physical space, whereas photos of an object often share only the object itself in many different contexts and times.
Figure~\ref{fig:teaser} shows an example of such photos, depicting the locomotive \emph{Flying Scotsman} captured over a century in many different locations.

Standard reconstruction pipelines are unreliable in such in-the-wild settings, and so are feedforward reconstruction models, which are largely trained on their output (Figure~\ref{fig:failure}).
Classical structure from motion relies on repeatable correspondences that agree with one rigid scene~\citep{schoenberger2016sfm}.
A changing environment provides few such correspondences, and the object may occupy only a small part of each photo.
Changes to an individual object over time and differences between physical units further weaken geometric consistency.
Together, these factors can result in incorrect reconstructions, so building a dataset from these photos requires identifying the photos of the intended object, matching them robustly, and verifying the recovered geometry.

In this work, we propose a dataset and an evaluation protocol for feedforward reconstruction of objects from in-the-wild photo collections.
To build this dataset, we collect candidate categories from Wikidata and Wikimedia Commons, such as \emph{Cristo Redentor}, the \emph{Hubble Space Telescope}, and the \emph{Ford Mustang}. 
We use a large language model (LLM) to retain the categories that depict a rigid physical object or design (Section~\ref{sec:dataset}).
To address the failures of standard pipelines on these collections, we propose a new object reconstruction pipeline, combining dense learned matching with geometric verification of the recovered camera poses and depth maps (Section~\ref{sec:pseudogt}).
The result of our pipeline is \datasetname{}, an in-the-wild object-centric dataset that contains 12{,}068 objects and 311{,}886 photos from Wikimedia Commons paired with camera poses, MVS depth, and object masks.
An example reconstruction of the \emph{Flying Scotsman} is shown on the right of Figure~\ref{fig:teaser}.

We evaluate recent feedforward reconstruction models on \datasetname{} and find that they perform substantially worse on these in-the-wild collections than on conventional benchmarks.
Therefore, we fine-tune one such model, VGGT-$\Omega$, on \datasetname{} and show that the resulting model, \modelname{}, significantly improves camera pose, depth, and 3D geometry on these objects, raising camera-pose AUC@30 from 25 to 81 on held-out objects and from 19 to 76 on NAVI in-the-wild~\citep{jampani2023navi}, an independent benchmark of objects photographed across different environments, while performance on conventional benchmarks is largely retained.
Together, these results suggest that \datasetname{} can help feedforward reconstruction models generalize to the real-world diversity of object photos.

%% file: sec/2_related.tex
\section{Related Work}
\label{sec:related}
\begin{figure*}[t]
\centering\includegraphics[width=\textwidth]{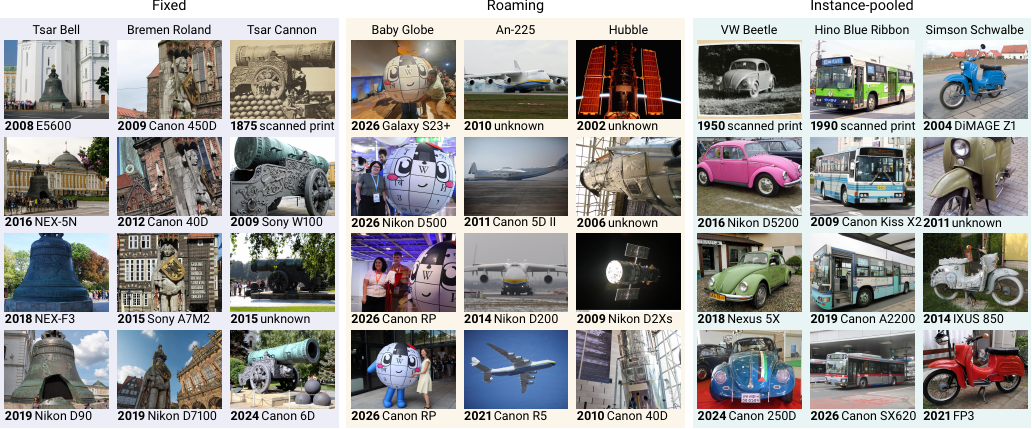}
\caption{\textbf{Examples of capture variation.}
Each column displays photos of example objects, grouped into the three regimes (fixed, roaming, and instance-pooled) and labeled with their year and camera.
\textit{Fixed} objects stay at one site, \textit{roaming} objects are single instances that move between sites, and \emph{instance-pooled} objects are different physical instances of one shared design.}
\label{fig:gallery}
\vspace{-0.75em}
\end{figure*}

\begin{table}[t]
\caption{\textbf{Real-image datasets for 3D reconstruction tasks.}
Datasets above the line record each object or scene in a single capture within minutes, while datasets below contain photos of each object or scene taken by many people spanning days to years.
NAVI and Pix3D align images to 3D models by hand; all other datasets recover poses with structure from motion.}
\label{tab:dataset-comparison}
\centering
\scriptsize
\setlength{\tabcolsep}{3pt}
\begin{tabular*}{\columnwidth}{@{\extracolsep{\fill}}lccccc@{}}
\toprule
Dataset & Type & Count & Images & Span & Depth \\
\midrule
OmniObject3D~\citep{wu2023omniobject3d} & objects & 6k & 1.2M & minutes & -- \\
CO3D~\citep{reizenstein2021co3d} & objects & 19k & 1.5M & minutes & MVS \\
MVImgNet2.0~\citep{han2024mvimgnet2} & objects & 520k & 15.5M & minutes & -- \\
uCO3D~\citep{liu2025uco3d} & objects & 170k & 34M & minutes & mono. \\
\midrule
NAVI~\citep{jampani2023navi} & objects & 36 & 10.5k & days & scan \\
Pix3D~\citep{sun2018pix3d} & objects & 395 & 10k & days/years & scan/CAD \\
MegaDepth~\citep{li2018megadepth} & scenes & 196 & 130k & years & MVS \\
MegaDepth-X~\citep{li2026megadepthx} & scenes & 1.9k & 440k & years & MVS \\
MegaScenes~\citep{tung2024megascenes} & scenes & 80k & 2M & years & -- \\
\datasetname{} (ours) & objects & 12k & 312k & years & MVS \\
\bottomrule
\end{tabular*}
\vspace{-1em}
\end{table}

\noindent\textbf{3D datasets.}
3D datasets have grown rapidly in scale.
Synthetic datasets such as Objaverse~\citep{deitke2023objaverse,deitke2023objaversexl} and ShapeNet~\citep{chang2015shapenet} provide textured 3D models.
Datasets built with 3D scanners include DTU~\citep{aanaes2016dtu}, Google Scanned Objects~\citep{downs2022gso}, and OmniObject3D~\citep{wu2023omniobject3d}, and contain at most a few thousand objects.
Meanwhile, CO3D~\citep{reizenstein2021co3d}, MVImgNet~\citep{yu2023mvimgnet,han2024mvimgnet2}, and uCO3D~\citep{liu2025uco3d} reach a larger scale by crowdsourcing videos of objects and annotating them with 3D information via structure from motion.
In such datasets, each object is recorded in one coordinated capture~\citep{reizenstein2021co3d,yu2023mvimgnet,liu2025uco3d,xia2024wildrgbd}.
ScanNet~\citep{dai2017scannet}, ScanNet++~\citep{yeshwanth2023scannetpp}, and DL3DV~\citep{ling2024dl3dv} record scenes in the same way.
Few datasets photograph objects in different environments.
PASCAL3D+~\citep{xiang2014pascal3d}, ObjectNet3D~\citep{xiang2016objectnet3d}, and Pix3D~\citep{sun2018pix3d} align CAD models to internet images of objects, but the models only approximate the depicted objects.
NAVI~\citep{jampani2023navi} is the closest dataset to our setting.
It photographs each object in different environments with different cameras and aligns each image to a 3D scan of that object.
In all of these object-scale datasets, each image is aligned to its corresponding 3D model by hand, and per-image geometry such as depth is rendered from that model rather than reconstructed from the photos.
In contrast, \datasetname{} consists of multi-view photo collections of objects taken by different people over many years, with camera poses and depth recovered by an automated reconstruction pipeline, and is much larger than NAVI.
Table~\ref{tab:dataset-comparison} compares \datasetname{} with other datasets of real images used for 3D reconstruction.

\medskip\noindent\textbf{Reconstruction from internet photo collections.}
Photo Tourism~\citep{snavely2006phototourism} and Building Rome in a Day~\citep{agarwal2009rome} reconstructed landmarks from unordered photos taken by many photographers.
Later datasets use such reconstructions as supervision for learned models.
MegaDepth~\citep{li2018megadepth} reconstructs about 200 landmarks from photos on Flickr with COLMAP~\citep{schoenberger2016sfm}, filters the resulting multi-view stereo depth, and uses it to supervise single-view depth prediction.
MegaScenes~\citep{tung2024megascenes} 
reconstructs 
80k Wikimedia Commons categories of scenes 
with COLMAP using 
SIFT features.
MegaDepth-X~\citep{li2026megadepthx} selects densely photographed MegaScenes scenes and reconstructs them with MASt3R-SfM~\citep{duisterhof2024mast3rsfm} and doppelganger disambiguation~\citep{xiangli2025doppelgangerspp}.
All three contain scene-level rather than object-centric reconstructions.
Of the 12{,}068 objects in \datasetname{}, only 1{,}163 (9.6\%) are reconstructed as scenes in MegaScenes and 56 (0.5\%) in MegaDepth-X, almost all of them objects permanently installed at one site, such as statues and memorials.
Objects that move between sites or exist as many physical units are almost entirely absent from these datasets.

\medskip\noindent\textbf{Learned matching and structure from motion.}
Classical structure from motion (SfM) methods like
COLMAP~\citep{schoenberger2016sfm}
match hand-crafted 
keypoints
and register images incrementally.
SuperPoint~\citep{detone2018superpoint} and SuperGlue~\citep{sarlin2020superglue} replace traditional keypoint detection and matching with learned components.
LoFTR~\citep{sun2021loftr} proposes detector-free matching, predicting matches directly with a transformer.
Dense matchers such as DKM~\citep{edstedt2023dkm}, RoMa~\citep{edstedt2024roma}, and RoMa v2~\citep{romav2} predict a warp for every pixel of an image pair, aiming to recover correspondences under large changes in viewpoint and illumination.
Several SfM systems, such as Detector-Free SfM~\citep{he2024dfsfm} and Dense-SfM~\citep{lee2025densesfm}, build on such dense matches to form multi-view tracks and refine the tracks together with the geometry.
MASt3R-SfM~\citep{duisterhof2024mast3rsfm} builds on the matches of MASt3R~\citep{leroy2024mast3r} and jointly aligns all images.
VGGSfM~\citep{wang2024vggsfm} learns the entire reconstruction pipeline, predicting tracks, cameras, and points and refining them with differentiable bundle adjustment.
We keep the incremental reconstruction of COLMAP and supply it with multi-view tracks built from RoMa v2 matches that pass confidence and geometric checks (Section~\ref{sec:pseudogt}).

\begin{figure*}[t]
\centering
\includegraphics[width=\textwidth]{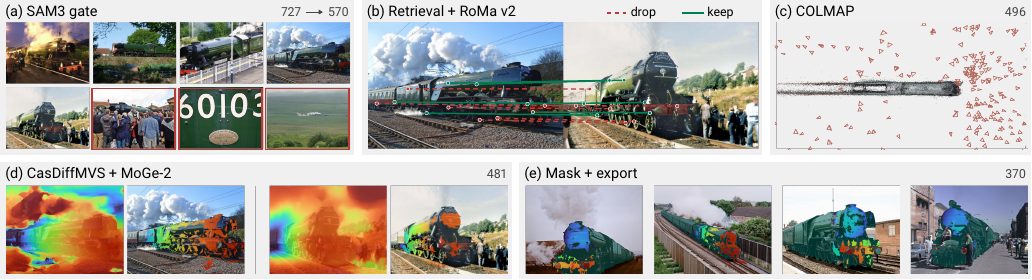}
\vspace{-10pt}
\caption{\textbf{Reconstruction pipeline for the Flying Scotsman.}
(a) Object gating retains 570 of 727 photos, with red borders marking excluded photos.
(b) A retrieved pair is shown with candidate correspondences.
Green correspondences pass the confidence, cycle, and epipolar checks, while dashed red correspondences fail.
(c) The structure-from-motion reconstruction registers 496 images, merged from 14 partial reconstructions.
View selection keeps 481 of them for multi-view stereo.
(d) The same pair shows MVS depth before and after MoGe-2 filtering.
(e) Four of the 370 photos in the dataset show MVS depth in color and the selected mask in teal over the image.}
\label{fig:pipeline}
\vspace{-0.75em}
\end{figure*}

\medskip\noindent\textbf{Feedforward 3D reconstruction.}
Feedforward models that regress cameras, depth, and point maps directly from a set of images have become a standard approach to 3D reconstruction~\citep{wang2024dust3r,leroy2024mast3r,yang2025fast3r,tang2025mvdust3r,zhang2025flare,wang2025vggt,wang2026pi3,lin2026da3,keetha2026mapanything,wang2026vggtomega}.
Most of their object-level training data come from single-session captures and renderings of synthetic 3D assets.
They are likewise evaluated on benchmarks built with such data.
Two recent works, MegaDepth-X~\citep{li2026megadepthx} and MegaUnScene~\citep{zhang2025megaunscene} train and evaluate feedforward models on in-the-wild internet photo collections, but only on scene-level, not object-centric, collections.
\datasetname{} is an object-level dataset similarly built from internet photos.
We fine-tune VGGT-$\Omega$ on this dataset and compare the result with recent feedforward models on \datasetname{} and NAVI in-the-wild (Section~\ref{sec:experiments}).

%% file: sec/3_dataset.tex
\section{The \datasetname{} Dataset}
\label{sec:dataset}

\datasetname{} is a dataset of objects reconstructed from Wikimedia Commons photos taken by many people over many years.
It contains many types of rigid objects (Table~\ref{tab:category-distribution}), including statues, cannons, pipe organs, telephone boxes, fountains, cameras, ships, aircraft, and locomotives.
This section describes how our dataset is curated from Internet sources, and Section~\ref{sec:pseudogt} describes how we reconstruct the individual objects.

\subsection{Objects from Wikimedia Commons}
\label{sec:dataset:identities}

Wikimedia Commons organizes photos into user-curated categories.
Categories for physical subjects are often specific to one object or design, such as \emph{Category:LNER Class A3 4472 Flying Scotsman}.
Each object in \datasetname{} corresponds to one such category, and we call the images filed under it a \emph{collection}.
Many of these categories are linked from Wikidata, the structured knowledge base supporting Wikimedia projects, and 68\% of the \datasetname{} objects have a Wikidata item.

Objects represent either individuals or designs.
An \emph{individual} is a single physical unit, such as one hull, one statue, or one numbered locomotive.
A \emph{design} is a shape shared by many physical units, such as a car model or a mass-produced telephone box with the same underlying geometry.
We pool all units of a design into one single collection, where individual units within the collection may differ in color, trim, and wear.

\subsection{Curation at scale}
\label{sec:dataset:curation}

We enumerate candidate categories in two ways.
Some classes, such as ships and locomotives, are enumerated from their Wikidata items; others are found by traversing the Wikimedia Commons category graph from root categories, such as \emph{Cameras} and \emph{Pipe organs}.
The traversal records 155{,}744 categories after excluding branches that do not depict the object itself, such as photos taken with a camera model, drawings, and replicas of an object.

The traversal also reaches subcategories for places, events, and parts related to an object, which we wish to exclude.
Title rules cannot reliably separate these categories from the objects themselves.
For example, \emph{Cannon Street station} is a railway station rather than a cannon.
Therefore, we use an LLM to classify each of the recorded categories from its title and the categories above it in the category graph.
The classifier assigns three labels: (i) whether the category depicts a rigid physical object, which decides whether it is kept; (ii) whether it names an individual or a design; and (iii) whether the object is permanently installed at one site.
Labels (ii) and (iii) determine the regime of each object (Section~\ref{sec:dataset:spectrum}).
Together with the Wikidata-enumerated classes, the categories kept by the classifier yield 33{,}884 candidate objects, of which we attempt to reconstruct 17{,}470 after filtering on the number of photos.

This curation alone does not filter individual images by their visual content.
Collections retain cluttered 
photos, partial views, and images in which the object occupies a small part of the frame.
We rely on the robust downstream reconstruction pipeline to either register these images in 3D or remove them when they cannot be registered correctly (Section~\ref{sec:pseudogt}).

\subsection{Capture variation}
\label{sec:dataset:spectrum}

\input{tables/category_distribution}

Different kinds of objects are characterized by different forms of variation in context and capture.
A statue fixed to its plinth yields photos that share a common environment, but differ in viewpoint, time, and lighting.
Photos of a locomotive span different stations with corresponding changes in the environment.
A telephone-box design represents many physical units whose exact wear and paint may differ.
We denote these distinct categories as \emph{fixed}, \emph{roaming}, and \emph{instance-pooled} 
(Figure~\ref{fig:gallery}).
\emph{Fixed} denotes an individual installed at one site, \emph{roaming} an individual that moves between sites, and \emph{instance-pooled} a design shared by many physical instances.
The labels follow from LLM classification (Section~\ref{sec:dataset:curation}) and are therefore approximate.

%% file: tables/category_distribution.tex
\begin{table}[t]
\caption{\textbf{Dataset composition.}
Number of objects in each supercategory.
An LLM groups the object labels assigned during curation into 53 categories and 13 supercategories.}
\label{tab:category-distribution}
\centering
\scriptsize
\begin{tabular*}{\columnwidth}{@{\extracolsep{\fill}}lrlr@{}}
\toprule
Supercategory & Objects & Supercategory & Objects \\
\midrule
Road vehicles & 3,341 & Household \& personal objects & 353 \\
Statues \& sculpture & 2,160 & Weapons \& armour & 342 \\
Monuments \& landmarks & 1,773 & Religious objects & 140 \\
Aircraft \& spacecraft & 1,461 & Street furniture & 130 \\
Rail vehicles & 1,108 & Machinery \& equipment & 121 \\
Watercraft & 701 & Other & 10 \\
Musical instruments & 428 & \textbf{Total} & \textbf{12,068} \\
\bottomrule
\end{tabular*}
\vspace{-1em}
\end{table}

%% file: sec/4_pseudogt.tex
\section{Pseudo-Ground Truth via Reconstruction}
\label{sec:pseudogt}

A Wikimedia Commons category can contain photos of other objects of the same kind, unrelated subjects, and crowded scenes in which the object is small.
In this section, we describe how we robustly reconstruct each object to obtain camera poses, MVS depth, and object masks via object gating, learned dense matching, and geometric verification.
Figure~\ref{fig:pipeline} follows one object through the reconstruction pipeline.

\subsection{Matching and structure from motion}
\label{sec:pseudogt:sfm}

\noindent\textbf{Object gating.}
We use an LLM to assign a text prompt to each object, similar to the classification in Section~\ref{sec:dataset:curation}.
We then use SAM3~\citep{sam3} with this prompt to identify images in which the object is the main subject.
We keep an image when the largest detected box covers at least 10\% of the image.
Common rejected images include crowd photos, close-ups of details, and images of a different subject filed in the same category.

\medskip\noindent\textbf{Dense matching.}
We compute point matches between photos with the dense matcher RoMa v2~\citep{romav2} due to its robustness to environment and object variation. 
In the supplement, we compare this choice with structure from motion from SIFT, SuperPoint with LightGlue, and MASt3R matches on the same images.
We embed each image with the retrieval model of MASt3R-SfM~\citep{duisterhof2024mast3rsfm} and match it against the 20 closest images in the embedding space.
For each image pair, RoMa v2 returns a dense mapping of every pixel of one image to the other along with a certainty map.
We sample candidate correspondences from these outputs.
We keep a correspondence between a point $\vx$ in one image and $W(\vx)$ in the other only if
\begin{equation}
  \begin{gathered}
  \sigma(\vx) \ge \tau_\sigma,
  \qquad
  \big\|W'\big(W(\vx)\big) - \vx\big\| \le \tau_{\mathrm{cyc}},
  \\
  d\big(W(\vx),\, \mF\vx\big) \le \tau_{\mathrm{epi}},
  \end{gathered}
  \label{eq:checks}
\end{equation}
where $W$ and $W'$ are the forward and backward warps, $\sigma$ is the certainty that RoMa v2 assigns to the match in both directions, and $\mF$ is a fundamental matrix estimated for the pair with MAGSAC++~\citep{barath2020magsacpp}.
The three conditions only retain confident, cycle-consistent matches that agree with the estimated epipolar geometry.
We merge nearby endpoints of the remaining correspondences into shared keypoints, following prior work on structure from motion from dense matches~\citep{he2022oneposepp,he2024dfsfm,lee2025densesfm}.
These shared keypoints link pairwise matches into multi-view tracks.

\medskip\noindent\textbf{Structure from motion.}
We import these keypoints and matches into COLMAP~\citep{schoenberger2016sfm} and perform incremental mapping.
We exhaustively merge reconstructed components and keep the largest result as the final reconstruction of the object.

\subsection{MVS depth and masks}
\label{sec:pseudogt:depth}
\begin{figure}[t]
\centering\includegraphics[width=\columnwidth]{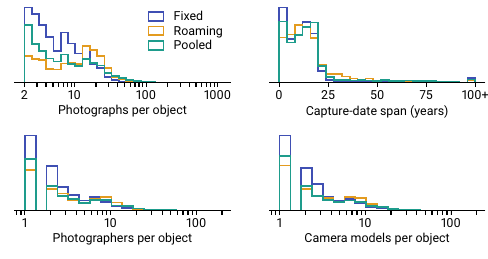}
\caption{\textbf{Object statistics by regime.} Distributions over \datasetname{} of the number of photos, the span of capture dates, the number of photographers, and the number of camera models.}
\label{fig:stats}
\vspace{-0.75em}
\end{figure}

\begin{figure*}[t]
\centering\includegraphics[width=\textwidth]{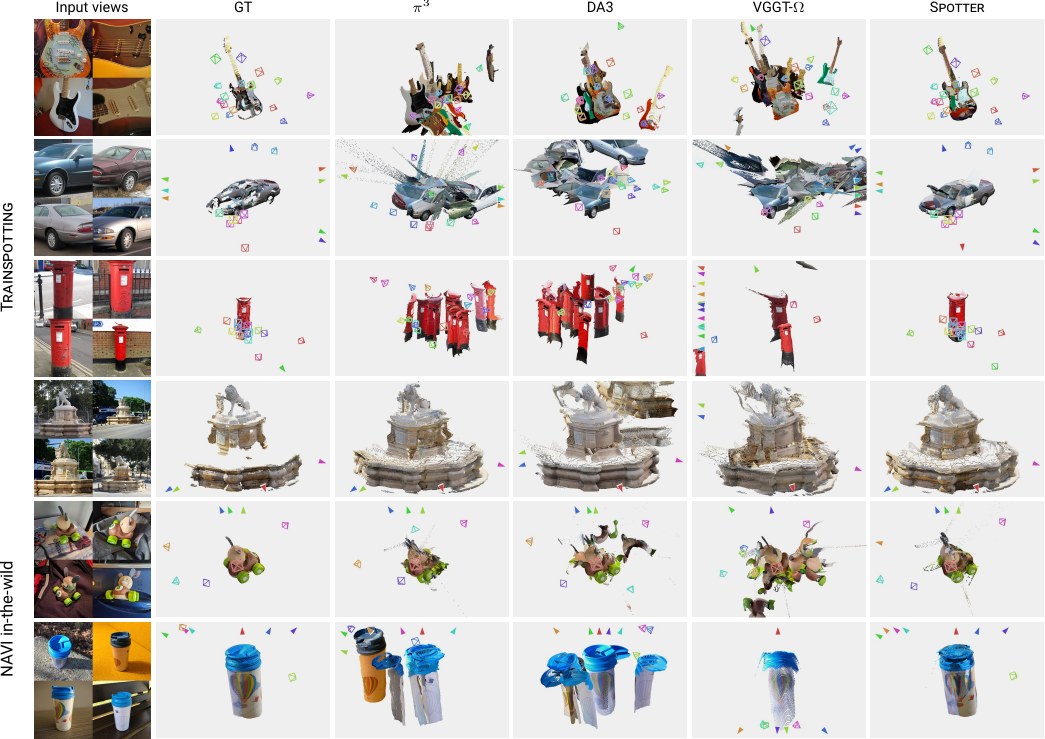}
\caption{\textbf{Qualitative comparison.}
We visualize ground truth and predicted camera poses and point maps from $\pi^3$~\citep{wang2026pi3}, Depth Anything 3~\citep{lin2026da3}, VGGT-$\Omega$~\citep{wang2026vggtomega}, and \modelname{} on held out objects from \datasetname{} and NAVI in-the-wild~\citep{jampani2023navi}. We use 4 or 16 input views for \datasetname{} and 8 for NAVI in-the-wild. Only the first 4 input views are shown. Triangular markers indicate the directions of the cameras outside of the frame.
The baselines misregister the input views and produce duplicated or fragmented geometry, whereas \modelname{} recovers a single consistent object and accurate camera poses.}
\label{fig:qualitative}
\vspace{-0.75em}
\end{figure*}

\noindent\textbf{View selection.}
A mis-registered view, or a view of a different object weakly linked to the reconstruction, is unreliable as ground truth and can corrupt the depth of its neighbors during MVS.
Therefore, we keep a view only if enough of its points have a track length of at least three.
We then build a covisibility graph in which two views are connected by an edge weighted by the number of 3D points they share.
We split this graph by spectral bisection and discard the smaller subgraph when the conductance of the cut is below 0.15.

\medskip\noindent\textbf{Multi-view stereo with a monocular check.}
We run Cas-DiffMVS~\citep{casdiffmvs} on the selected views and keep high-confidence depths.
Since MVS can be unreliable on textureless and specular surfaces, we verify these depths against a monocular depth prediction from MoGe-2~\citep{moge2}, as in MegaDepth-X~\citep{li2026megadepthx}, using the spherical filtering strategy of MoGe-2 and ViGeo~\citep{moge2,vigeo}. 

\medskip\noindent\textbf{Object masks.}
We select the SAM3 mask that covers the largest share of the verified depth as the target object mask.
A photo is included only if enough of the selected mask carries verified depth.
A final quality check removes objects with erroneous reconstructions and photos with implausible focal lengths or in which the object is not the dominant subject.

\subsection{Statistics}
\label{sec:pseudogt:stats}

We pass 17{,}470 objects with 1{,}251{,}986 photos through our reconstruction pipeline.
We report the number of objects and photos that remain after each stage in the supplement.
Object gating removes the largest share of photos.
Structure from motion succeeds for 85\% of the objects and registers 72\% of the gated photos.

\medskip\noindent\textbf{Dataset.}
Our final dataset, \datasetname{}, contains 12{,}068 objects and 311{,}886 photos.
Out of all of its objects, 4{,}338 are fixed, 2{,}869 roaming, and 4{,}861 instance-pooled.
Capture dates are available for 84\% of the photos. The median capture span is 12 years and 90\% of the objects have a capture span of at least one year.
A median object has photos by 7 different photographers taken with 6 different camera models.
Figure~\ref{fig:stats} shows these distributions by regime.
For a coarse view of the composition, we use an LLM to group the 337 object labels of the dataset into 53 categories and 13 supercategories (Table~\ref{tab:category-distribution}).
We release \datasetname{} with its camera poses, MVS depth, and object masks.
Every photo is distributed with its Wikimedia Commons source, author, and licence.

\medskip\noindent\textbf{Split.}
We partition \datasetname{} by object into a training set of 11{,}068 objects and 279{,}319 photos and a held-out set of 1{,}000 objects and 32{,}567 photos, so that no photo of a held-out object is seen during training.

%% file: sec/6_experiments.tex
\section{Experiments}
\label{sec:experiments}

\subsection{Setup}
\label{sec:experiments:setup}

\noindent\textbf{Model and training.}
We obtain \modelname{} by fine-tuning the released 1B-parameter checkpoint of VGGT-$\Omega$~\citep{wang2026vggtomega} on the training objects of \datasetname{}.
We update the camera head and all attention blocks of the transformer backbone.
Each training sample contains one object and between 2 and 24 of its photos, chosen so that they form a connected set under covisibility.
We consider two photos covisible when at least 5\% of the object points of one project inside the object mask of the other.

\medskip\noindent\textbf{Objective.}
Following VGGT, we express the cameras and depths of each sample in the coordinate frame of its first view.
We normalize the ground truth by the mean distance of its valid 3D points to the origin.
We normalize the predictions in the same way, following DUSt3R~\citep{wang2024dust3r}.
The model is trained with a combined camera, depth, and point loss:
\begin{align}
  \mathcal{L} &= 5\,\mathcal{L}_{\text{cam}} + \mathcal{L}_{\text{depth}} + \mathcal{L}_{\text{point}},
  \label{eq:loss}\\
  \mathcal{L}_{\text{cam}} &= \textstyle\sum_{i} \big\|\hat{\vg}_i - \vg_i\big\|_1,
  \label{eq:loss-cam}\\
  \mathcal{L}_{\text{depth}} &= \tfrac{1}{|\mathcal{V}|}\textstyle\sum_{\vx\in\mathcal{V}} \big(1+\tfrac{1}{D(\vx)}\big)\,\big|\hat{D}(\vx)-D(\vx)\big|,
  \label{eq:loss-depth}\\
  \mathcal{L}_{\text{point}} &= \tfrac{1}{|\mathcal{V}|}\textstyle\sum_{\vx\in\mathcal{V}} \big\|\hat{\vp}(\vx)-\vp(\vx)\big\|_2,
  \label{eq:loss-point}
\end{align}
where $\hat{\vg}_i$ is the predicted camera of view $i$, encoded as translation, rotation quaternion, and field of view, $\mathcal{V}$ is the set of valid pixels, and $\hat{\vp}(\vx)$ is the 3D point of pixel $\vx$ given by the predicted depth and camera.
For the depth and point losses, we use plain regression loss without confidence weighting and discard the largest 2\% of residuals in each batch.

\medskip\noindent\textbf{Replay.}
To limit forgetting, a fixed 15\% of every batch contains samples from BlendedMVS~\citep{yao2020blendedmvs}, Hypersim~\citep{roberts2021hypersim}, and uCO3D~\citep{liu2025uco3d}, supervised with the same loss.

\medskip\noindent\textbf{Object masks.}
Since the background differs between the photos of an in-the-wild collection, the model also predicts an object mask for each view, with a small head trained on the dataset masks on top of the frozen features.
This allows restricting the point map output to the object, separating it from its background.
The head is trained in a separate stage with the rest of the model frozen and does not affect any of the evaluations below.
Details are described in the supplement.

\medskip\noindent\textbf{Implementation details.}
We train with AdamW~\citep{loshchilov2019adamw} at a peak learning rate of $10^{-5}$ and a weight decay of $0.05$.
We use linear warm-up over the first 5\% of steps and a cosine decay.
Training runs for 7,500 steps, with 768 images in each step, and takes about 12 hours on four NVIDIA B200 GPUs.

\medskip\noindent\textbf{Baselines.}
We compare \modelname{} with the released checkpoints of VGGT~\citep{wang2025vggt}, $\pi^3$~\citep{wang2026pi3}, Depth Anything 3~\citep{lin2026da3}, MapAnything~\citep{keetha2026mapanything}, and VGGT-$\Omega$~\citep{wang2026vggtomega}.

\subsection{Evaluation protocol}
\label{sec:experiments:protocol}

\noindent\textbf{Samples.}
On \datasetname{}, we draw one connected 4-view sample from each held-out object and, when the object has enough photos, one connected 16-view sample, for 1{,}635 samples from 1{,}000 objects.
On each external dataset, we use one sample of 8 views per scene, equally spaced along the capture sequence when one exists and drawn at random otherwise, since published protocols differ in the number and selection of input views.
Samples are drawn with a fixed seed.

\medskip\noindent\textbf{Camera pose.}
We compare the predicted and ground truth cameras on every pair of views in a sample.
For each pair, we compute the error of the relative rotation and the direction of the relative translation and report the area under the accuracy curve of the larger of the two errors up to $5^\circ$ and $30^\circ$ (AUC@5 and AUC@30), following PoseDiffusion~\citep{wang2023posediffusion}.
All metrics are computed per sample and then averaged over samples.

\medskip\noindent\textbf{Depth.}
We evaluate the predicted depth maps on the pixels with ground-truth depth.
For models that only predict point maps, we take depth as the z-coordinate of the point map in each camera's frame.
We scale the predicted depth of all views in a sample by a single median ratio to the ground truth.
We report the absolute relative error (AbsRel) and the inlier ratio $\delta<1.03$, the fraction of pixels whose predicted and ground-truth depths differ by less than a factor of 1.03.

\input{tables/object_stack}
\input{tables/conventional}
\medskip\noindent\textbf{Point maps.}
For models that only predict depth, we obtain a point map for each view by unprojecting the predicted depth with the predicted camera.
For models that predict point maps, we use the predicted point maps directly.
We align the predicted points to the ground-truth points with a similarity transform estimated from corresponding pixels and refined by iterative closest point (ICP).
We then report accuracy, the mean distance from each predicted point to the nearest ground-truth point, and completeness, the mean distance from each ground-truth point to the nearest predicted point, following Spann3R~\citep{wang2025spann3r}.
On \datasetname{}, whose reconstructions have an arbitrary scale, we express both in percent of the object extent; on the external datasets, we express them in millimeters.

\subsection{Results on \datasetname{}}
\label{sec:experiments:main}
Table~\ref{tab:object-main} reports evaluation results on \datasetname{}.
\modelname{} improves on VGGT-$\Omega$ in every metric, most strongly in camera pose and completeness, and substantially outperforms the strongest baseline, $\pi^3$.
Table~\ref{tab:object-breakdown} shows that the difficulty is concentrated in roaming and instance-pooled objects, whose photos differ most in environment.
VGGT-$\Omega$ performs considerably better on fixed objects than on the other two regimes, whereas \modelname{} performs similarly on all three, and performance further improves with more views.
Figure~\ref{fig:qualitative} shows reconstructions of four \datasetname{} objects and two NAVI in-the-wild objects.
The baselines fail to register views taken in different environments.
Their point maps therefore contain duplicated or fragmented copies of the object, and their cameras are scattered.
\modelname{} registers the same views consistently and reconstructs a single object.

\subsection{NAVI in-the-wild}
\label{sec:experiments:navi}

The NAVI dataset~\citep{jampani2023navi} contains 36 objects, each with a scanned mesh and camera poses aligned to it, photographed both in \emph{multiview} captures recorded in a single session and in an \emph{in-the-wild} collection across different environments.
As the held-out \datasetname{} results are scored against pseudo-ground truth from our own pipeline, evaluating on NAVI in-the-wild provides an independent measure of accuracy.
Under our protocol NAVI yields 35 in-the-wild and 281 multiview samples.
Tables~\ref{tab:conventional-geom} and~\ref{tab:conventional-camera} report the results.
VGGT-$\Omega$ reaches an AUC@30 of 91.6 on the multiview captures but only 18.8 on the in-the-wild collections of the same objects.
Fine-tuning on \datasetname{} raises the in-the-wild AUC@30 to 76.0 and improves completeness from 20.4 to 1.9\,mm.
\modelname{} achieves the best in-the-wild result on every metric except accuracy, on which all models lie within 1\,mm of one another.
We note that when the views of a prediction are misregistered, the similarity alignment scales the prediction down toward the center of the object, so each predicted point stays close to the surface while most of the surface is left uncovered.
Pose failures therefore appear as completeness error rather than accuracy error.
On the multiview captures, performance is largely unchanged.

\subsection{Conventional benchmarks}
\label{sec:experiments:conventional}

We evaluate all models on 7-Scenes~\citep{shotton2013sevenscenes}, Neural RGB-D~\citep{azinovic2022nrgbd}, DTU~\citep{aanaes2016dtu}, and ETH3D~\citep{schops2017eth3d} to measure the effect of fine-tuning on conventional data (Tables~\ref{tab:conventional-geom} and~\ref{tab:conventional-camera}).
Relative to VGGT-$\Omega$, fine-tuning changes AUC@30 by at most 1.1 points and $\delta<1.03$ by at most 3.2 points on these four datasets.

\subsection{Ablations}
\label{sec:experiments:ablations}
Table~\ref{tab:ablation-data} compares models trained on subsets of the \datasetname{} training objects.
All models in this comparison, including the reference model trained on all objects, are trained for 10 epochs with the same recipe as \modelname{}.
We find that the types of photos used for training matter more than their scale.
Training only on fixed objects yields lower accuracy than training on a random subset half its size, both on held-out objects and on NAVI in-the-wild.
It is also considerably less accurate on roaming and instance-pooled objects, whereas models trained on those regimes remain close to the reference model when evaluated on fixed objects.
This shows that photos across different environments, which fixed objects and existing datasets largely lack, are necessary for generalization to the other regimes and to NAVI in-the-wild.

\input{tables/ablation_data}

%% file: tables/object_stack.tex
\begin{table}[t]
\caption{\textbf{Results on \datasetname{}.} Point-map accuracy and completeness are given as a percentage of the object extent after a similarity alignment refined by ICP. Depth AbsRel and inlier ratio $\delta_{1.03}$ are measured after a single median scale per sample. Camera-pose AUC at $5^\circ$ and $30^\circ$ is computed over all pairs of views (Section~\ref{sec:experiments:protocol}).}
\label{tab:object-main}
\centering\scriptsize\setlength{\tabcolsep}{3pt}
\begin{tabularx}{\columnwidth}{l*{6}{>{\centering\arraybackslash}X}}
\toprule
Method & \multicolumn{2}{c}{Point map} & \multicolumn{2}{c}{Depth} & \multicolumn{2}{c}{Camera} \\
\cmidrule(lr){2-3} \cmidrule(lr){4-5} \cmidrule(lr){6-7}
 & Acc$\downarrow$ & Comp$\downarrow$ & AbsRel$\downarrow$ & $\delta_{1.03}\uparrow$ & AUC$_{5}\uparrow$ & AUC$_{30}\uparrow$ \\
\midrule
VGGT~\citep{wang2025vggt} & \cellcolor[HTML]{FDD2D3}2.38 & \cellcolor[HTML]{FEE8BF}8.08 & \cellcolor[HTML]{FEE2C1}0.160 & \cellcolor[HTML]{FDD9C7}32.4 & \cellcolor[HTML]{FDD6CA}11.5 & \cellcolor[HTML]{FEE2C1}44.1 \\
$\pi^3$~\citep{wang2026pi3} & \cellcolor[HTML]{FDD2D0}2.33 & \cellcolor[HTML]{F5F1BB}6.22 & \cellcolor[HTML]{FEE1C2}0.162 & \cellcolor[HTML]{FEDFC3}34.7 & \cellcolor[HTML]{FDD7C9}12.5 & \cellcolor[HTML]{FEE6C0}46.4 \\
DA3~\citep{lin2026da3} & \cellcolor[HTML]{FDD5CB}2.22 & \cellcolor[HTML]{FDD2D0}12.17 & \cellcolor[HTML]{FDD4CC}0.189 & \cellcolor[HTML]{FDD8C8}32.1 & \cellcolor[HTML]{FDD4CC}10.1 & \cellcolor[HTML]{FDD3CE}29.4 \\
MapAny.~\citep{keetha2026mapanything} & \cellcolor[HTML]{FDD6CA}2.20 & \cellcolor[HTML]{FEEBBE}7.75 & \cellcolor[HTML]{FDD2D3}0.206 & \cellcolor[HTML]{FDD2D3}26.6 & \cellcolor[HTML]{FDD2D3}5.3 & \cellcolor[HTML]{FDD5CB}33.1 \\
VGGT-$\Omega$~\citep{wang2026vggtomega} & \cellcolor[HTML]{FFF0BD}1.84 & \cellcolor[HTML]{FDD2D3}12.68 & \cellcolor[HTML]{FDD2D0}0.201 & \cellcolor[HTML]{FDD3CF}28.1 & \cellcolor[HTML]{FDD2D0}7.2 & \cellcolor[HTML]{FDD2D3}25.0 \\
\modelname{} & \cellcolor[HTML]{A9DBB6}\textbf{1.22} & \cellcolor[HTML]{A9DBB6}\textbf{1.15} & \cellcolor[HTML]{A9DBB6}\textbf{0.071} & \cellcolor[HTML]{A9DBB6}\textbf{53.3} & \cellcolor[HTML]{A9DBB6}\textbf{44.2} & \cellcolor[HTML]{A9DBB6}\textbf{81.2} \\
\bottomrule
\end{tabularx}
\par\vspace{10pt}
\caption{\textbf{Results by view count and regime.} AUC@30 on the held-out \datasetname{} objects. Each \datasetname{} object contributes one 4-view sample and, if it has at least 16 photos, one 16-view sample. The regime columns pool both sample sizes.}
\label{tab:object-breakdown}
\centering\scriptsize
\begin{tabularx}{\columnwidth}{l*{5}{>{\centering\arraybackslash}X}}
\toprule
Method & 4 views & 16 views & Fixed & Roaming & Pooled \\
 & (1000) & (635) & (442) & (450) & (743) \\
\midrule
VGGT-$\Omega$~\citep{wang2026vggtomega} & \cellcolor[HTML]{FDD3CE}29.4 & \cellcolor[HTML]{FDD2D3}18.1 & \cellcolor[HTML]{F4F1BB}56.7 & \cellcolor[HTML]{FDD2D3}14.9 & \cellcolor[HTML]{FDD2D3}12.3 \\
\modelname{} & \cellcolor[HTML]{ADDCB7}\textbf{79.9} & \cellcolor[HTML]{A9DBB6}\textbf{83.2} & \cellcolor[HTML]{B0DDB7}\textbf{79.0} & \cellcolor[HTML]{AADBB6}\textbf{81.0} & \cellcolor[HTML]{A9DBB6}\textbf{82.6} \\
\bottomrule
\end{tabularx}
\vspace{-1em}
\end{table}

%% file: tables/conventional.tex
\begin{table*}[t]
\input{tables/conventional_geom}
\par\vspace{10pt}
\input{tables/conventional_camera}
\end{table*}

%% file: tables/conventional_geom.tex
\caption{\textbf{Geometry on external benchmarks.} Point-map accuracy and completeness are given in millimeters after a similarity alignment refined by ICP. Depth inlier ratio $\delta_{1.03}$ is measured after a single median scale per sample. Every model is given the same 8 views of each scene. NAVI photographs each object in one session (\emph{NAVI}) and across different environments (\emph{NAVI (wild)}). The other datasets record each scene in a single capture.}
\label{tab:conventional-geom}
\centering\scriptsize\setlength{\tabcolsep}{0.5pt}
\begin{tabularx}{\textwidth}{l*{18}{>{\centering\arraybackslash}X}}
\toprule
Method & \multicolumn{3}{c}{NAVI (wild)} & \multicolumn{3}{c}{NAVI} & \multicolumn{3}{c}{7-Scenes} & \multicolumn{3}{c}{NRGBD} & \multicolumn{3}{c}{DTU} & \multicolumn{3}{c}{ETH3D} \\
\cmidrule(lr){2-4} \cmidrule(lr){5-7} \cmidrule(lr){8-10} \cmidrule(lr){11-13} \cmidrule(lr){14-16} \cmidrule(lr){17-19}
 & Acc$\downarrow$ & Comp$\downarrow$ & $\delta_{1.03}\uparrow$ & Acc$\downarrow$ & Comp$\downarrow$ & $\delta_{1.03}\uparrow$ & Acc$\downarrow$ & Comp$\downarrow$ & $\delta_{1.03}\uparrow$ & Acc$\downarrow$ & Comp$\downarrow$ & $\delta_{1.03}\uparrow$ & Acc$\downarrow$ & Comp$\downarrow$ & $\delta_{1.03}\uparrow$ & Acc$\downarrow$ & Comp$\downarrow$ & $\delta_{1.03}\uparrow$ \\
\midrule
VGGT~\citep{wang2025vggt} & \cellcolor[HTML]{BAE0B7}3.2 & \cellcolor[HTML]{FFF2BC}11.5 & \cellcolor[HTML]{FEE8BF}35.0 & \cellcolor[HTML]{C9E5B8}2.5 & \cellcolor[HTML]{C2E2B8}3.4 & \cellcolor[HTML]{E7EDBA}71.5 & \cellcolor[HTML]{FEE3C1}25.3 & \cellcolor[HTML]{FEEABE}32.6 & \cellcolor[HTML]{F7F2BB}67.5 & \cellcolor[HTML]{C3E3B8}17.4 & \cellcolor[HTML]{EDEFBA}16.9 & \cellcolor[HTML]{CDE6B8}90.9 & \cellcolor[HTML]{A9DBB6}\textbf{1.2} & \cellcolor[HTML]{FEDBC5}1.8 & \cellcolor[HTML]{A9DBB6}\textbf{97.4} & \cellcolor[HTML]{FEEBBE}291.4 & \cellcolor[HTML]{FFEFBD}317.5 & \cellcolor[HTML]{FDD2D1}52.6 \\
$\pi^3$~\citep{wang2026pi3} & \cellcolor[HTML]{FEE8BF}3.7 & \cellcolor[HTML]{F1F0BB}9.6 & \cellcolor[HTML]{C1E2B8}41.2 & \cellcolor[HTML]{A9DBB6}\textbf{2.2} & \cellcolor[HTML]{A9DBB6}\textbf{3.0} & \cellcolor[HTML]{C3E3B8}76.5 & \cellcolor[HTML]{A9DBB6}\textbf{20.0} & \cellcolor[HTML]{A9DBB6}\textbf{26.7} & \cellcolor[HTML]{D8E9B9}70.5 & \cellcolor[HTML]{C6E4B8}17.9 & \cellcolor[HTML]{EFEFBA}17.2 & \cellcolor[HTML]{DCEAB9}88.6 & \cellcolor[HTML]{A9DBB6}\textbf{1.2} & \cellcolor[HTML]{A9DBB6}\textbf{1.5} & \cellcolor[HTML]{BBE0B7}91.2 & \cellcolor[HTML]{D0E6B9}154.7 & \cellcolor[HTML]{C4E3B8}129.7 & \cellcolor[HTML]{FEE1C2}62.4 \\
DA3~\citep{lin2026da3} & \cellcolor[HTML]{CCE5B8}3.3 & \cellcolor[HTML]{FDDAC6}16.0 & \cellcolor[HTML]{C3E3B8}41.1 & \cellcolor[HTML]{FEE0C3}3.3 & \cellcolor[HTML]{FDD2D1}5.7 & \cellcolor[HTML]{A9DBB6}\textbf{80.0} & \cellcolor[HTML]{FEE2C1}25.4 & \cellcolor[HTML]{FDD4CC}35.7 & \cellcolor[HTML]{B3DEB7}74.1 & \cellcolor[HTML]{BDE1B8}16.4 & \cellcolor[HTML]{E0EBBA}15.3 & \cellcolor[HTML]{A9DBB6}\textbf{96.7} & \cellcolor[HTML]{F1F0BB}1.7 & \cellcolor[HTML]{D4E8B9}1.6 & \cellcolor[HTML]{ACDCB7}96.5 & \cellcolor[HTML]{FFF0BC}274.3 & \cellcolor[HTML]{F5F1BB}270.3 & \cellcolor[HTML]{E1EBBA}73.2 \\
MapAny.~\citep{keetha2026mapanything} & \cellcolor[HTML]{FDD2D3}4.1 & \cellcolor[HTML]{FEF4BB}11.0 & \cellcolor[HTML]{FDD2D3}29.6 & \cellcolor[HTML]{FDD2D3}3.8 & \cellcolor[HTML]{FDD2D3}5.8 & \cellcolor[HTML]{FDD2D3}56.4 & \cellcolor[HTML]{FDD2D3}28.2 & \cellcolor[HTML]{FDD2D3}36.8 & \cellcolor[HTML]{FDD2D3}58.3 & \cellcolor[HTML]{FDD2D3}42.8 & \cellcolor[HTML]{FDD2D3}30.2 & \cellcolor[HTML]{FDD2D3}69.2 & \cellcolor[HTML]{FDD2D3}2.4 & \cellcolor[HTML]{FDD2D3}1.9 & \cellcolor[HTML]{FDD2D3}37.8 & \cellcolor[HTML]{FDD2D3}459.7 & \cellcolor[HTML]{FDD2D3}542.0 & \cellcolor[HTML]{FDD2D3}51.9 \\
VGGT-$\Omega$~\citep{wang2026vggtomega} & \cellcolor[HTML]{A9DBB6}\textbf{3.1} & \cellcolor[HTML]{FDD2D3}20.4 & \cellcolor[HTML]{FDD2D2}29.7 & \cellcolor[HTML]{F4F1BB}2.9 & \cellcolor[HTML]{FDD7C9}5.3 & \cellcolor[HTML]{C6E3B8}76.1 & \cellcolor[HTML]{CFE6B9}21.8 & \cellcolor[HTML]{B5DFB7}27.4 & \cellcolor[HTML]{CAE5B8}71.9 & \cellcolor[HTML]{ABDCB7}13.3 & \cellcolor[HTML]{AFDDB7}9.0 & \cellcolor[HTML]{C4E3B8}92.4 & \cellcolor[HTML]{A9DBB6}\textbf{1.2} & \cellcolor[HTML]{FFF4BB}1.7 & \cellcolor[HTML]{AADBB6}97.1 & \cellcolor[HTML]{ADDCB7}73.6 & \cellcolor[HTML]{ACDCB7}61.1 & \cellcolor[HTML]{A9DBB6}\textbf{83.5} \\
\modelname{} & \cellcolor[HTML]{BAE0B7}3.2 & \cellcolor[HTML]{A9DBB6}\textbf{1.9} & \cellcolor[HTML]{A9DBB6}\textbf{43.1} & \cellcolor[HTML]{FFECBE}3.1 & \cellcolor[HTML]{E6EDBA}4.0 & \cellcolor[HTML]{D1E7B9}74.5 & \cellcolor[HTML]{ADDCB7}20.2 & \cellcolor[HTML]{B5DFB7}27.4 & \cellcolor[HTML]{A9DBB6}\textbf{75.1} & \cellcolor[HTML]{A9DBB6}\textbf{12.9} & \cellcolor[HTML]{A9DBB6}\textbf{8.3} & \cellcolor[HTML]{C2E2B8}92.7 & \cellcolor[HTML]{A9DBB6}\textbf{1.2} & \cellcolor[HTML]{D4E8B9}1.6 & \cellcolor[HTML]{AEDCB7}95.9 & \cellcolor[HTML]{A9DBB6}\textbf{65.5} & \cellcolor[HTML]{A9DBB6}\textbf{52.9} & \cellcolor[HTML]{ADDCB7}82.8 \\
\bottomrule
\end{tabularx}
\vspace{-0.5em}

%% file: tables/conventional_camera.tex
\caption{\textbf{Camera pose on external benchmarks.} Camera-pose AUC at $5^\circ$ and $30^\circ$ is computed over all pairs of the same 8 views of each scene.}
\label{tab:conventional-camera}
\centering\scriptsize\setlength{\tabcolsep}{1.5pt}
\begin{tabularx}{\textwidth}{l*{12}{>{\centering\arraybackslash}X}}
\toprule
Method & \multicolumn{2}{c}{NAVI (wild)} & \multicolumn{2}{c}{NAVI} & \multicolumn{2}{c}{7-Scenes} & \multicolumn{2}{c}{NRGBD} & \multicolumn{2}{c}{DTU} & \multicolumn{2}{c}{ETH3D} \\
\cmidrule(lr){2-3} \cmidrule(lr){4-5} \cmidrule(lr){6-7} \cmidrule(lr){8-9} \cmidrule(lr){10-11} \cmidrule(lr){12-13}
 & AUC@5$\uparrow$ & AUC@30$\uparrow$ & AUC@5$\uparrow$ & AUC@30$\uparrow$ & AUC@5$\uparrow$ & AUC@30$\uparrow$ & AUC@5$\uparrow$ & AUC@30$\uparrow$ & AUC@5$\uparrow$ & AUC@30$\uparrow$ & AUC@5$\uparrow$ & AUC@30$\uparrow$ \\
\midrule
VGGT~\citep{wang2025vggt} & \cellcolor[HTML]{FEE9BF}14.4 & \cellcolor[HTML]{FFF0BD}45.5 & \cellcolor[HTML]{C0E2B8}53.3 & \cellcolor[HTML]{B8E0B7}91.2 & \cellcolor[HTML]{FEE5C0}26.6 & \cellcolor[HTML]{F1F0BB}81.0 & \cellcolor[HTML]{C2E2B8}86.4 & \cellcolor[HTML]{BFE2B8}97.6 & \cellcolor[HTML]{A9DBB6}\textbf{98.1} & \cellcolor[HTML]{A9DBB6}\textbf{99.7} & \cellcolor[HTML]{FDD4CC}37.6 & \cellcolor[HTML]{FDD8C7}71.1 \\
$\pi^3$~\citep{wang2026pi3} & \cellcolor[HTML]{FEE4C1}13.1 & \cellcolor[HTML]{FFECBE}43.7 & \cellcolor[HTML]{B2DEB7}55.8 & \cellcolor[HTML]{B0DDB7}91.6 & \cellcolor[HTML]{FFF0BC}28.8 & \cellcolor[HTML]{D6E8B9}83.3 & \cellcolor[HTML]{C1E2B8}86.5 & \cellcolor[HTML]{C8E4B8}97.2 & \cellcolor[HTML]{CFE6B9}80.0 & \cellcolor[HTML]{BDE1B8}96.6 & \cellcolor[HTML]{FDD3CE}36.4 & \cellcolor[HTML]{FEDDC4}72.9 \\
DA3~\citep{lin2026da3} & \cellcolor[HTML]{FEE0C2}12.0 & \cellcolor[HTML]{FEE0C2}36.9 & \cellcolor[HTML]{A9DBB6}\textbf{57.2} & \cellcolor[HTML]{ABDCB7}91.8 & \cellcolor[HTML]{FFEEBD}28.5 & \cellcolor[HTML]{EBEEBA}81.5 & \cellcolor[HTML]{A9DBB6}\textbf{91.4} & \cellcolor[HTML]{A9DBB6}\textbf{98.6} & \cellcolor[HTML]{B8DFB7}91.0 & \cellcolor[HTML]{B1DEB7}98.5 & \cellcolor[HTML]{E5ECBA}55.4 & \cellcolor[HTML]{FEE5C0}75.4 \\
MapAny.~\citep{keetha2026mapanything} & \cellcolor[HTML]{FDD2D3}3.3 & \cellcolor[HTML]{FDD3CD}24.2 & \cellcolor[HTML]{FDD2D3}28.4 & \cellcolor[HTML]{FDD2D3}84.0 & \cellcolor[HTML]{FDD2D3}18.2 & \cellcolor[HTML]{FDD2D3}72.4 & \cellcolor[HTML]{FDD2D3}56.5 & \cellcolor[HTML]{FDD2D3}90.9 & \cellcolor[HTML]{FDD2D3}15.0 & \cellcolor[HTML]{FDD2D3}73.4 & \cellcolor[HTML]{FDD2D3}33.8 & \cellcolor[HTML]{FDD2D3}65.6 \\
VGGT-$\Omega$~\citep{wang2026vggtomega} & \cellcolor[HTML]{FDD2D2}3.8 & \cellcolor[HTML]{FDD2D3}18.8 & \cellcolor[HTML]{B4DEB7}55.4 & \cellcolor[HTML]{B0DDB7}91.6 & \cellcolor[HTML]{A9DBB6}\textbf{40.8} & \cellcolor[HTML]{A9DBB6}\textbf{87.2} & \cellcolor[HTML]{BBE0B7}87.8 & \cellcolor[HTML]{B9E0B7}97.9 & \cellcolor[HTML]{B8DFB7}91.1 & \cellcolor[HTML]{B1DEB7}98.5 & \cellcolor[HTML]{A9DBB6}\textbf{66.9} & \cellcolor[HTML]{A9DBB6}\textbf{91.8} \\
\modelname{} & \cellcolor[HTML]{A9DBB6}\textbf{30.7} & \cellcolor[HTML]{A9DBB6}\textbf{76.0} & \cellcolor[HTML]{ACDCB7}56.7 & \cellcolor[HTML]{A9DBB6}\textbf{91.9} & \cellcolor[HTML]{B8E0B7}38.8 & \cellcolor[HTML]{B6DFB7}86.1 & \cellcolor[HTML]{C1E2B8}86.5 & \cellcolor[HTML]{C6E4B8}97.3 & \cellcolor[HTML]{C2E2B8}86.1 & \cellcolor[HTML]{B6DFB7}97.7 & \cellcolor[HTML]{B1DDB7}65.4 & \cellcolor[HTML]{A9DBB6}\textbf{91.8} \\
\bottomrule
\end{tabularx}
\vspace{-0.5em}

%% file: tables/ablation_data.tex
\begin{table}[t]
\caption{\textbf{Training-data ablations.} AUC@30 of models trained for 10 epochs on subsets of the \datasetname{} training objects. The number of training objects is given in parentheses.}
\label{tab:ablation-data}
\centering\scriptsize\setlength{\tabcolsep}{3pt}
\begin{tabularx}{\columnwidth}{l*{4}{>{\centering\arraybackslash}X}c}
\toprule
 & \multicolumn{4}{c}{\datasetname{} held-out} & \\
\cmidrule(lr){2-5}
Training data & All & Fixed & Roaming & Pooled & NAVI (wild) \\
\midrule
All (11{,}068) & \cellcolor[HTML]{B6DFB7}80.3 & \cellcolor[HTML]{C5E3B8}78.2 & \cellcolor[HTML]{BCE1B7}79.5 & \cellcolor[HTML]{A9DBB6}82.1 & \cellcolor[HTML]{A9DBB6}74.6 \\
Random 20\% (2{,}193) & \cellcolor[HTML]{BAE0B7}79.7 & \cellcolor[HTML]{CAE5B8}77.5 & \cellcolor[HTML]{BEE1B8}79.2 & \cellcolor[HTML]{AFDDB7}81.3 & \cellcolor[HTML]{CCE5B8}72.3 \\
Fixed (4{,}038) & \cellcolor[HTML]{FFECBE}68.5 & \cellcolor[HTML]{C4E3B8}78.4 & \cellcolor[HTML]{FDD2D3}58.0 & \cellcolor[HTML]{FFEEBD}68.9 & \cellcolor[HTML]{FDD2D3}63.4 \\
Roaming (2{,}603) & \cellcolor[HTML]{BFE2B8}79.0 & \cellcolor[HTML]{D1E7B9}76.5 & \cellcolor[HTML]{B6DFB7}80.3 & \cellcolor[HTML]{BAE0B7}79.7 & \cellcolor[HTML]{FDD8C8}65.7 \\
Pooled (4{,}427) & \cellcolor[HTML]{C1E2B8}78.8 & \cellcolor[HTML]{D7E9B9}75.6 & \cellcolor[HTML]{CEE6B8}77.0 & \cellcolor[HTML]{ACDCB7}81.7 & \cellcolor[HTML]{B4DEB7}73.9 \\
\bottomrule
\end{tabularx}
\vspace{-1em}
\end{table}

%% file: sec/8_conclusion.tex
\section{Conclusion}
\label{sec:conclusion}

Current feedforward reconstruction models perform poorly on in-the-wild object-centric photo collections, even for objects that they reconstruct accurately from a single capture session.
Such in-the-wild collections are abundant in public photo archives, and \datasetname{} shows that they can be turned into training data at scale.
Fine-tuning VGGT-$\Omega$ on \datasetname{} improves camera pose, depth, and 3D geometry on held-out objects and on the independent NAVI in-the-wild benchmark, while performance on conventional benchmarks remains largely unchanged.
The base models had already been trained on large datasets of objects recorded in single capture sessions, and we also show that training on fixed objects alone transfers poorly to roaming and instance-pooled objects, so what these models lacked was variation in how an object is photographed.
Beyond training, \datasetname{} provides a benchmark for reconstruction under changing environments, and its reconstructions of objects photographed over decades allow studying how objects and their surroundings change over time.

%% file: sec/9_acknowledgments.tex
\section*{Acknowledgments}
This material is based upon work supported by the Naval Air Warfare Center under Contract No.\ N68335-25-C-0482.
We gratefully acknowledge use of the research computing resources of the Empire AI Consortium, Inc, with support from Empire State Development of the State of New York, the Simons Foundation, and the Secunda Family Foundation.

%% file: sec/S1_curation.tex
\section{Dataset Curation}
\label{sec:supp:curation}

\input{tables/supp_categories}
This section gives the details of the curation summarized in Section~3.2 of the main paper.

\medskip\noindent\textbf{Enumeration.}
Thirteen vehicle classes are enumerated from their Wikidata items: aircraft, ships, locomotives, automobiles, rail vehicles, buses, trams, military vehicles, trucks, motorcycles, tractors, construction equipment, and other vehicles.
Each item gives one Wikimedia Commons category, 22{,}983 in total.
Whether a class names individual objects or designs is fixed by the class: aircraft, ships, locomotives, and trams are identified by registration, hull, or fleet number, while the remaining classes are identified by model.

All other objects are found by traversing the Wikimedia Commons category graph from 856 root categories.
We chose 211 of them by hand, and the LLM screened the remaining 645 from the subcategories of broad hand-chosen categories.
The traversal does not descend into branches whose titles match one of seven kinds of category that do not depict the object itself: photos taken with a camera model, people with an object, media about an object, replicas and scale models, institutions, parts, and wiki maintenance categories.
It prunes 37{,}930 branches this way and records 155{,}744 categories.

\medskip\noindent\textbf{Classification.}
Of the recorded categories, 49{,}252 contain at least 25 photos, counting the photos of subcategories that depict the same object.
The LLM classifies each of them from its title and the titles of the categories above it in the graph.
The prompt is reproduced in Section~\ref{sec:supp:prompts}.
It keeps 40{,}254 categories as rigid physical objects, rejects 8{,}956 for the reasons listed in Table~\ref{tab:supp-reject}, and leaves 42 undecided.
Of the kept categories, 13{,}347 name an individual object or a design, and the remaining 26{,}907 are categories of many objects, such as \emph{Statues in Paris}, which we do not use.
After removing categories already covered by the vehicle classes and merging subcategories that depict the same object into their parent, 10{,}901 object categories remain.
Together with the vehicle classes, they form the 33{,}884 candidate objects of the main paper.

\begin{table}[H]
\caption{\textbf{Categories rejected by the classifier.} Reasons for the 8{,}956 categories with at least 25 photos that the LLM does not keep.}
\label{tab:supp-reject}
\centering\scriptsize
\begin{tabular*}{\columnwidth}{@{\extracolsep{\fill}}lr@{}}
\toprule
Reason & Categories \\
\midrule
Flat object, such as a painting or a sign & 1{,}802 \\
Media or two-dimensional depiction of an object & 1{,}606 \\
Place or institution & 1{,}489 \\
Abstract or maintenance category & 1{,}066 \\
Event or activity & 811 \\
Deformable object & 767 \\
Part of an object & 614 \\
People & 275 \\
Photos taken with or of a product & 246 \\
Other & 280 \\
\midrule
Total & 8{,}956 \\
\bottomrule
\end{tabular*}
\end{table}

\medskip\noindent\textbf{Categories.}
Table~\ref{tab:supp-categories} lists the 53 categories into which the LLM groups the 337 object labels, with the number of objects and photos of each.

\medskip\noindent\textbf{Overlap with internet-photo scene datasets.}
Table~\ref{tab:supp-overlap} gives the overlap of Section~2 of the main paper by regime.
A category counts as a MegaScenes scene when its title matches a category collected by MegaScenes, and as reconstructed when MegaScenes releases a reconstruction for it.
Nearly all overlapping objects are fixed objects, which MegaScenes and MegaDepth-X treat as scenes.

\input{tables/supp_overlap}

%% file: tables/supp_categories.tex
\begin{table*}[t]
\caption{\textbf{Object categories.} The 53 categories, grouped by supercategory, with the number of objects and photos of each (12{,}068 objects and 311{,}886 photos in total).}
\label{tab:supp-categories}
\centering\scriptsize
\begin{tabular*}{\textwidth}{@{\extracolsep{\fill}}lrr@{\hspace{3em}}lrr@{}}
\toprule
Category & Objects & Photos & Category & Objects & Photos \\
\midrule
\multicolumn{3}{l}{\textit{Road vehicles}} & \multicolumn{3}{l}{\textit{Musical instruments}} \\
\quad car & 1,967 & 87,193 & \quad pipe organ & 390 & 5,733 \\
\quad bus & 476 & 28,932 & \quad musical instrument & 38 & 982 \\
\quad military vehicle & 341 & 7,522 & \multicolumn{3}{l}{\textit{Household \& personal objects}} \\
\quad truck \& van & 269 & 8,553 & \quad camera \& lens & 142 & 2,946 \\
\quad motorcycle & 187 & 2,776 & \quad computer \& office machine & 90 & 1,364 \\
\quad tractor \& other road vehicle & 71 & 522 & \quad vase \& vessel & 47 & 893 \\
\quad bicycle & 30 & 506 & \quad consumer electronics & 33 & 457 \\
\multicolumn{3}{l}{\textit{Statues \& sculpture}} & \quad appliance \& personal item & 24 & 260 \\
\quad statue & 1,628 & 24,940 & \quad furniture \& lighting & 17 & 178 \\
\quad sculpture & 432 & 7,649 & \multicolumn{3}{l}{\textit{Weapons \& armour}} \\
\quad bust \& head & 100 & 1,576 & \quad artillery & 173 & 2,088 \\
\multicolumn{3}{l}{\textit{Monuments \& landmarks}} & \quad firearm \& hand weapon & 127 & 1,550 \\
\quad war memorial & 553 & 8,079 & \quad missile & 29 & 286 \\
\quad fountain & 385 & 4,788 & \quad helmet & 13 & 144 \\
\quad monument \& memorial & 340 & 8,276 & \multicolumn{3}{l}{\textit{Religious objects}} \\
\quad obelisk & 141 & 2,205 & \quad altar \& altarpiece & 68 & 504 \\
\quad gravestone \& tomb & 76 & 1,173 & \quad church furnishing \& reliquary & 30 & 291 \\
\quad megalith & 73 & 1,131 & \quad cross \& crucifix & 25 & 261 \\
\quad sarcophagus \& coffin & 46 & 685 & \quad bell & 17 & 261 \\
\quad door, gate \& arch & 45 & 949 & \multicolumn{3}{l}{\textit{Street furniture}} \\
\quad clock \& sundial & 39 & 1,005 & \quad post box & 85 & 3,003 \\
\quad runestone \& carved stone & 33 & 657 & \quad bench \& other street furniture & 28 & 325 \\
\quad commemorative column \& pillar & 31 & 302 & \quad telephone box & 17 & 733 \\
\quad tower \& small building & 11 & 182 & \multicolumn{3}{l}{\textit{Machinery \& equipment}} \\
\multicolumn{3}{l}{\textit{Aircraft \& spacecraft}} & \quad engine & 62 & 618 \\
\quad aircraft & 1,415 & 37,798 & \quad industrial machine \& equipment & 24 & 385 \\
\quad spacecraft \& rocket & 46 & 647 & \quad crane & 14 & 135 \\
\multicolumn{3}{l}{\textit{Rail vehicles}} & \quad scientific instrument & 12 & 171 \\
\quad locomotive & 543 & 16,838 & \quad robot & 9 & 111 \\
\quad tram & 496 & 16,033 & \multicolumn{3}{l}{\textit{Other}} \\
\quad train \& railcar & 69 & 2,493 & \quad other object & 10 & 61 \\
\multicolumn{3}{l}{\textit{Watercraft}} & & & \\
\quad ship \& boat & 701 & 14,736 & & & \\
\bottomrule
\end{tabular*}
\end{table*}

%% file: tables/supp_overlap.tex
\begin{table}[H]
\caption{\textbf{Overlap with internet-photo scene datasets by regime.} Number of \datasetname{} objects whose Wikimedia Commons category is a MegaScenes scene, has a released MegaScenes reconstruction, or is a MegaDepth-X scene.}
\label{tab:supp-overlap}
\centering\scriptsize
\begin{tabular*}{\columnwidth}{@{\extracolsep{\fill}}lrrrr@{}}
\toprule
Regime & Objects & MegaScenes & \multicolumn{1}{c}{MegaScenes} & MegaDepth-X \\
 & & category & reconstruction & \\
\midrule
Fixed & 4,338 & 1,798 & 1,128 & 56 \\
Roaming & 2,869 & 41 & 34 & 0 \\
Instance-pooled & 4,861 & 2 & 1 & 0 \\
\midrule
Total & 12,068 & 1,841 & 1,163 & 56 \\
\bottomrule
\end{tabular*}
\end{table}

%% file: sec/S2_pipeline.tex
\section{Reconstruction Pipeline}
\label{sec:supp:pipeline}

Table~\ref{tab:recon-yield} lists the objects and photos that remain after each stage of the pipeline.
Table~\ref{tab:supp-thresholds} lists the fixed thresholds of the pipeline of Section~4 of the main paper.

\begin{table}[t]
\caption{\textbf{Reconstruction yield.}
Number of objects and photos that remain after each stage of the pipeline in Section~4 of the main paper.}
\label{tab:recon-yield}
\centering
\scriptsize
\begin{tabular*}{\columnwidth}{@{\extracolsep{\fill}}lrr@{}}
\toprule
Stage & Objects & Photos \\
\midrule
Input & 17{,}470 & 1{,}251{,}986 \\
Object gating & 16{,}141 & 775{,}609 \\
Structure from motion & 14{,}818 & 558{,}281 \\
MVS depth & 14{,}032 & 486{,}275 \\
Mask and depth checks & 13{,}236 & 367{,}245 \\
Final dataset & 12{,}068 & 311{,}886 \\
\bottomrule
\end{tabular*}
\end{table}

\begin{figure*}[t]
\centering\includegraphics[width=\textwidth]{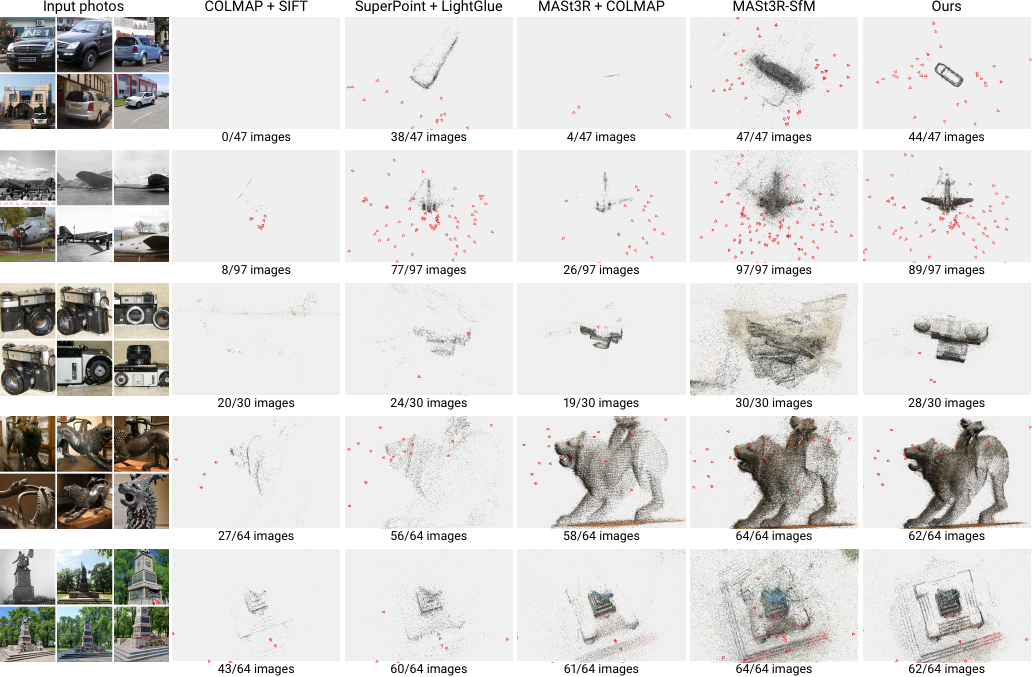}
\caption{\textbf{Structure from motion with other matchers.}
Each row shows one \datasetname{} object: six of the input photos, then the largest model of COLMAP with SIFT, SuperPoint with LightGlue, MASt3R matches with the COLMAP mapper, MASt3R-SfM, and our pipeline, all run on the same images.
The number of registered images out of the input images is given under each panel.
The other methods fail on instance-pooled and roaming objects, and on fixed objects our pipeline produces a more complete and cleaner reconstruction.
In the bottom three rows, we frame the object to show its details, so not all cameras are visible.}
\label{fig:supp-sfm-baselines}
\end{figure*}

\medskip\noindent\textbf{Comparison with other matchers.}
Figure~\ref{fig:supp-sfm-baselines} compares the pipeline of Section~4 of the main paper with four alternatives on the images that pass object gating: COLMAP with SIFT and exhaustive matching~\citep{lowe2004sift,schoenberger2016sfm}, SuperPoint with LightGlue~\citep{detone2018superpoint,lindenberger2023lightglue} in hloc~\citep{sarlin2019coarse} with NetVLAD retrieval~\citep{arandjelovic2016netvlad}, MASt3R matches~\citep{leroy2024mast3r} with the COLMAP mapper, and MASt3R-SfM~\citep{duisterhof2024mast3rsfm}.
Each alternative runs with its default settings and, where it retrieves image pairs, with 20 neighbours per image as in our pipeline.
On the instance-pooled and roaming objects, the alternatives fail to reconstruct the object.
Even on the fixed objects, our pipeline produces a more complete and cleaner reconstruction.

%% file: sec/S3_training.tex
\section{Training Details}
\label{sec:supp:training}

\input{tables/supp_thresholds}

This section completes the setup of Section~5.1 of the main paper.

\medskip\noindent\textbf{Trainable parameters.}
We fine-tune the released VGGT-$\Omega$ checkpoint with 1B parameters.
All 24 frame and global attention layers of the aggregator and the camera head are trained, while the patch embedding, the camera and register tokens, and the depth head are frozen.
This leaves 808M of the 1.14B parameters trainable.

\medskip\noindent\textbf{Sampling.}
The number of views per sample is drawn once per batch, uniformly between 2 and 24.
Each batch is filled to a fixed budget of 192 images per GPU.
Inputs follow the training preprocessing of VGGT-$\Omega$~\citep{wang2026vggtomega}.
Depth and points are supervised only on object pixels that have MVS depth.
The replay samples from BlendedMVS, Hypersim, and uCO3D are supervised on all pixels with ground truth.

\medskip\noindent\textbf{Optimization.}
The cosine schedule decays the learning rate to $10^{-8}$.
Gradients are clipped to a norm of 1.
Training uses bfloat16 precision with gradient checkpointing.

\medskip\noindent\textbf{Object mask head.}
After fine-tuning, we freeze the model and train a small convolutional head on the fused features of its depth head to predict, for each pixel, whether it belongs to the object.
The target is the object mask of the dataset.
On held-out objects, the predicted masks reach an intersection over union of 0.92 with the dataset masks.

%% file: tables/supp_thresholds.tex
\begin{table*}[t]
\caption{\textbf{Pipeline thresholds.} The fixed thresholds of the reconstruction pipeline of Section~4 of the main paper, in the order of its stages. Image-relative values are fractions of the image side.}
\label{tab:supp-thresholds}
\centering\scriptsize\setlength{\tabcolsep}{5pt}
\begin{tabularx}{\textwidth}{lllX}
\toprule
Stage & Threshold & Value & Definition \\
\midrule
Object gating & minimum box size & 10\% of the image area & Largest SAM3 detection, with detection score above 0.5. \\
Retrieval & neighbours per image & 20 &  \\
Matching & correspondences per pair & 8{,}000 & Sampled from the dense warp. \\
 & $\tau_\sigma$ & 0.20 & Certainty threshold. \\
 & $\tau_{\mathrm{cyc}}$ & 0.4\% of the image side & Cycle-consistency distance. \\
 & $\tau_{\mathrm{epi}}$ & 0.1\% of the image side & Epipolar distance. \\
 & keypoint merge radius & 0.2\% of the image side & Match endpoints within this radius form one keypoint. \\
View selection & confirmed points & 100 & Points on tracks of length at least 3. \\
 & confirmed fraction & 0.30 & Confirmed points over all triangulated points of the image. \\
 & covisibility cut & conductance below 0.15 & Spectral bisection of the covisibility graph; the largest cluster is kept. \\
Multi-view stereo & source views & 6 &  \\
 & confidence & 0.3, 0.5, 0.5 & Minimum confidence at the three stages of CasDiffMVS. \\
Mask selection & confident depth & 200 px & Minimum confident MVS pixels for a mask to be chosen. \\
 & verified depth & 5\% of the mask & Minimum fraction of the chosen mask with verified depth. \\
Final quality check & dominant subject & $5\times$ & Chosen mask area relative to every other detection. \\
 & focal length & 0.3 to 10 image widths &  \\
 & photos per object & 2 (training), 5 (held-out) &  \\
\bottomrule
\end{tabularx}
\end{table*}

%% file: sec/S4_evaluation.tex
\section{Evaluation Details}
\label{sec:supp:evaluation}

This section details the protocol of Section~5.2 of the main paper.

\medskip\noindent\textbf{Samples on \datasetname{}.}
The views of a sample form a connected set under the covisibility relation used for training: two photos are connected when at least 5\% of the object points of one project inside the object mask of the other.
Each view is resized to cover $512\times512$ pixels and cropped around its principal point.

\medskip\noindent\textbf{External datasets.}
For ordered captures, the 8 views are equally spaced over the sequence, including its first and last frame.
For unordered collections, they are a fixed random draw.
Scenes with fewer than 8 usable frames are skipped.
On NAVI, images are center-cropped to a square, and frames that the dataset flags as occluded are excluded.
Inputs are resized to each model's patch multiple, and outputs are resampled to the ground-truth grid.

\medskip\noindent\textbf{Metrics.}
The depth scale of a sample is the median ratio of ground-truth to predicted depth over the valid pixels of all its views.
The similarity transform of the point-map metrics is estimated from corresponding valid pixels~\citep{umeyama1991least} before the ICP refinement.

\medskip\noindent\textbf{Effect of the alignment on misregistered predictions.}
Section~5.4 of the main paper explains why pose failures appear as completeness error rather than accuracy error.
Table~\ref{tab:supp-extent} quantifies this on NAVI in-the-wild by the extent of the aligned prediction relative to the extent of the ground truth.
The alignment scales most predictions of VGGT-$\Omega$ to less than half the object's size, while the predictions of \modelname{} keep their size.

\begin{table}[H]
\caption{\textbf{Extent of the aligned prediction on NAVI in-the-wild.} Ratio of the extent of the aligned prediction to the extent of the ground truth over the 35 objects, and the number of objects with a ratio below 0.7.}
\label{tab:supp-extent}
\centering\scriptsize
\begin{tabular*}{\columnwidth}{@{\extracolsep{\fill}}lcccc@{}}
\toprule
Method & Median & 10th pct. & 90th pct. & Objects below 0.7 \\
\midrule
VGGT-$\Omega$ & 0.43 & 0.15 & 0.87 & 28 \\
\modelname{} & 0.92 & 0.81 & 0.96 & 0 \\
\bottomrule
\end{tabular*}
\end{table}

%% file: sec/S6_qualitative.tex
\section{Additional Qualitative Results}
\label{sec:supp:qualitative}

\begin{figure}[t]
\centering\includegraphics[width=\columnwidth]{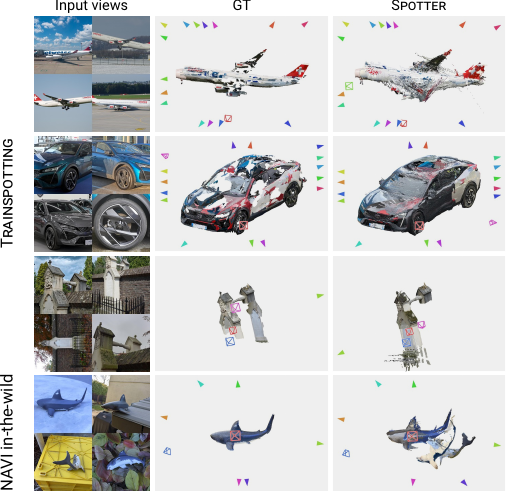}
\caption{\textbf{Failure cases.}
Four held-out samples on which \modelname{} fails: airliner (noisy depth at long focal lengths), car (misregistered close-up of a wheel), grave monument (doppelganger ambiguity), and toy shark from NAVI in-the-wild (out of distribution).}
\label{fig:supp-failures}
\end{figure}

\medskip\noindent\textbf{Failure cases.}
Figure~\ref{fig:supp-failures} shows four held-out samples on which \modelname{} fails, one for each of the most common failure modes.
Views taken with long focal lengths are ambiguous between focal length and depth, so their points can be misaligned with the other views and appear as noise around the airliner.
Close-up views can be ambiguous in their placement on the object, and the close-up of a wheel is localized at the wrong position on the car.
Symmetric objects produce doppelganger ambiguities, and the cameras of the grave monument are placed on the wrong one of its two identical pillars.
Objects far from the training distribution, such as the toy shark, whose appearance changes drastically with its pose and placement across the views, are reconstructed inconsistently.

\medskip\noindent\textbf{Additional qualitative comparisons.}
Figure~\ref{fig:supp-random} shows ten held-out samples drawn at random.

%% file: sec/S5_results.tex
\section{Additional Results}
\label{sec:supp:results}

Table~\ref{tab:supp-object-full} reports all metrics of every model on \datasetname{}, and Table~\ref{tab:supp-external-full} on the external benchmarks.

%% file: sec/S7_prompts.tex
\section{LLM Prompts}
\label{sec:supp:prompts}

All three LLM passes of the curation use the same setup.
The model is Claude Opus 5, the system prompt is the rubric reproduced below, each user message contains a batch of 50 numbered categories, and the response is JSON constrained to a fixed schema.
On a hand-labelled set of 32 categories, the classification pass agreed with the labels on all 32 verdicts and on 14 of 15 granularity labels.

\medskip\noindent\textbf{Category classification} (Section~3.2 of the main paper).
The response for each category gives the verdict (keep, reject, or uncertain), the reject reason, the granularity (individual, model, named unknown, or pool), and whether the object is installed at a fixed site.
\par\smallskip\noindent\rule{\columnwidth}{0.3pt}\par\nobreak\smallskip
\begingroup\scriptsize\ttfamily\RaggedRight\setlength{\parindent}{0pt}\setlength{\parskip}{0pt}
You are classifying Wikimedia Commons category titles for a 3D object reconstruction dataset. Each category{\char39}s images will be used to reconstruct a rigid physical object with COLMAP. Your job is SUBJECT identity: decide from the title (plus its parent category and crawl root) whether the category is about physical objects suitable for reconstruction.\par
\smallskip
Each numbered item gives:\par
\hangindent=1.2em\hangafter=1 - the category TITLE (your main evidence),\par
\hangindent=1.2em\hangafter=1 - parent: the category it was found under,\par
\hangindent=1.2em\hangafter=1 - root: the top-level category a breadth-first crawl started from to reach it. A hint only -- the crawl sometimes wanders, so an item may sit under an unrelated root. An off-root category of genuine physical objects is still a keep.\par
\hangindent=1.2em\hangafter=1 - files: how many image files the category holds,\par
\hangindent=1.2em\hangafter=1 - wikidata: whether a Wikidata item is linked to it. A weak hint that the category names a recognized concept -- but concepts include generic types ({\char34}settee{\char34}), so it proves nothing about specificity.\par
\smallskip
The pipeline tolerates noise but not wrongness: a rejected category is gone forever, while a wrongly kept one merely wastes a download before reconstruction discards it. When a title does not clearly establish a reject reason, KEEP it.\par
\smallskip
For each numbered item return:\par
\smallskip
verdict:\par
\hangindent=1.2em\hangafter=1 - {\char34}keep{\char34} -- the category depicts physical, rigid objects (whatever family they belong to).\par
\hangindent=1.2em\hangafter=1 - {\char34}reject{\char34} -- the title alone establishes one of the reject reasons below.\par
\hangindent=1.2em\hangafter=1 - {\char34}uncertain{\char34} -- genuinely cannot tell what the category depicts.\par
\smallskip
reason (for reject; use {\char34}ok{\char34} when keep):\par
\hangindent=1.2em\hangafter=1 - {\char34}usage\_or\_provenance{\char34} -- photos taken WITH the object or sourced from somewhere: {\char34}Taken with Nikon D5300{\char34}, {\char34}Photos from X stream{\char34}, {\char34}Media from iNaturalist{\char34}.\par
\hangindent=1.2em\hangafter=1 - {\char34}people{\char34} -- people using/holding/wearing the object: {\char34}Guitarists with...{\char34}, {\char34}People with Canon cameras{\char34}.\par
\hangindent=1.2em\hangafter=1 - {\char34}media\_or\_2d{\char34} -- depictions rather than the object: logos, icons, SVG, diagrams, drawings, paintings, stamps, advertisements, book titles, close-up detail shots.\par
\hangindent=1.2em\hangafter=1 - {\char34}place\_or\_institution{\char34} -- a place, street, building, company, museum, maker, or event venue that merely contains the family noun: {\char34}Cannon Street station{\char34}, {\char34}Scientific instrument makers{\char34}.\par
\hangindent=1.2em\hangafter=1 - {\char34}event\_or\_activity{\char34} -- competitions, festivals, ceremonies, activities: {\char34}2019 ICF Canoe Slalom World Championships{\char34}, {\char34}Emptying of post boxes{\char34}.\par
\hangindent=1.2em\hangafter=1 - {\char34}abstract\_or\_meta{\char34} -- licenses, wiki maintenance, pronunciation audio, quality-image collections, categories-about-categories.\par
\hangindent=1.2em\hangafter=1 - {\char34}flat{\char34} -- near-planar subjects: manhole covers, coins, plaques, murals, signage, banknotes.\par
\hangindent=1.2em\hangafter=1 - {\char34}deformable\_or\_soft{\char34} -- clothing, uniforms, textiles, bags, plush.\par
\hangindent=1.2em\hangafter=1 - {\char34}parts\_or\_accessories{\char34} -- components rather than whole objects: {\char34}Guitar headstocks{\char34}, {\char34}Camera parts{\char34}, {\char34}Bicycle pannier racks{\char34} is a borderline whole object (keep); use this only for clear fragments of a larger object.\par
\hangindent=1.2em\hangafter=1 - {\char34}other{\char34} -- a reject that fits none of the above (say nothing else; the verdict is what matters).\par
\smallskip
granularity (for keep; {\char34}n\_a{\char34} for reject):\par
\hangindent=1.2em\hangafter=1 - {\char34}individual{\char34} -- one specific physical object: {\char34}Penfold pillar box on Otterndorf Green, Sheringham{\char34}, {\char34}USCGC White Pine (WLM-547){\char34}.\par
\hangindent=1.2em\hangafter=1 - {\char34}model{\char34} -- one manufactured design or standardized type whose units are geometrically interchangeable: {\char34}Fender Stratocaster{\char34}, {\char34}K6 telephone boxes in Derbyshire{\char34}, {\char34}Queen Elizabeth II type A/B red pillar boxes in Merseyside{\char34}, {\char34}IBM Model M Keyboard{\char34}. A place qualifier does not demote a model.\par
\hangindent=1.2em\hangafter=1 - {\char34}named\_unknown{\char34} -- the title carries a proper-name or model-number token you do not recognize, modifying the family noun: {\char34}Parc Vue benches in New York City{\char34}, {\char34}KX100 telephone boxes{\char34}. An unfamiliar name is evidence of specificity -- never downgrade these to pool just because you don{\char39}t know the name.\par
\hangindent=1.2em\hangafter=1 - {\char34}pool{\char34} -- a generic type, possibly place-scoped, containing many different designs: {\char34}Snowplows{\char34}, {\char34}Settees{\char34}, {\char34}Traffic bollards in Wuppertal{\char34}, {\char34}Cannons in Florida{\char34}, {\char34}Wooden benches{\char34}. Also under-specified family names that span multiple geometries: {\char34}Glock{\char34} (17/19/26/43 differ), {\char34}King George V post boxes in the Republic of Ireland{\char34} (mixes wall-mounted and pillar).\par
\smallskip
fixed\_location (boolean): true if the objects are permanently installed at fixed sites (street furniture, installed fountains, monuments-like); false if portable/movable; judge from the object family when the title doesn{\char39}t say.\par
\smallskip
object\_scale (boolean, for keep; use false for reject): a descriptive FLAG, never a reason to reject. True if the subject is hand-to-room scale -- a statue, gravestone, cannon, telescope on a mount, church organ. False if larger. The test: could a person walk around it and photograph the whole thing from a few metres? {\char34}Very Large Telescope{\char34} (an observatory site) -\textgreater{} false; {\char34}Zenit avtomat{\char34} (a camera) -\textgreater{} true. Large MOVABLE things -- ships, airliners, locomotives -- are false and that is fine; the flag records size only, and such categories are still kept. Obelisks and triumphal columns -\textgreater{} true; anything with interior rooms -\textgreater{} false.\par
\smallskip
Different PHYSICAL UNITS of one design count as one object. A category like {\char34}Honda Accord Crosstour{\char34} contains photos of many distinct physical cars -- different owners, colors, countries -- yet it is ONE object for reconstruction, because every unit shares the same body shape. Likewise fifty K6 telephone kiosks across fifty towns are one {\char34}model{\char34}. A category is never a pool merely because its photos show more than one physical specimen.\par
\smallskip
The line between {\char34}model{\char34} and {\char34}pool{\char34} is SHAPE, not appearance. Reconstruction only uses geometry: units that differ in color, finish, livery, trim, or wear but share one body shape are ONE model ({\char34}Fender Stratocaster{\char34} stays a model despite countless colors; {\char34}Red post boxes{\char34} pooled across designs is still a pool despite one color). A category is a pool only when the objects in it have materially different SHAPES -- different products, different size variants (Glock 17 vs 26), or a bare type noun spanning many designs. Minor sub-variant drift (a changed headstock logo, a facelifted grille) does not demote a model; a doubt about shape goes to {\char34}model{\char34}, not {\char34}pool{\char34}.\par
\smallskip
Rules of thumb:\par
\hangindent=1.2em\hangafter=1 - Distinguish {\char34}type + place{\char34} pools from {\char34}named design + place{\char34} models by whether the pre-place part names a specific design (code, reign+type, brand+model) or a bare noun.\par
\hangindent=1.2em\hangafter=1 - A brand alone ({\char34}Zenit cameras{\char34}) is a pool of that brand{\char39}s models -- keep, pool.\par
\hangindent=1.2em\hangafter=1 - Museum/display context does not reject an object category: {\char34}Torpedoes{\char34} in museums are still torpedoes. Only reject when the subject is the institution itself.\par
\hangindent=1.2em\hangafter=1 - When two readings are plausible, keep it and give your best-guess granularity -- do not resolve doubt by rejecting.\par
\endgroup\par\nobreak\smallskip\noindent\rule{\columnwidth}{0.3pt}\par\medskip
\noindent User message:
\par\smallskip\noindent\rule{\columnwidth}{0.3pt}\par\nobreak\smallskip
\begingroup\scriptsize\ttfamily\RaggedRight\setlength{\parindent}{0pt}\setlength{\parskip}{0pt}
Classify these Wikimedia Commons categories per the rubric:\par
\smallskip
0. {\char34}\textless{}title\textgreater{}{\char34} | parent: \textless{}parent category\textgreater{} | root: \textless{}crawl root\textgreater{} | files: \textless{}n\textgreater{} | wikidata: yes/no\par
1. ...\par
\endgroup\par\nobreak\smallskip\noindent\rule{\columnwidth}{0.3pt}\par\medskip

\medskip\noindent\textbf{Segmentation prompt} (Section~4.1 of the main paper).
The response for each category gives one generic noun phrase, optional alternatives, and a confidence.
\par\smallskip\noindent\rule{\columnwidth}{0.3pt}\par\nobreak\smallskip
\begingroup\scriptsize\ttfamily\RaggedRight\setlength{\parindent}{0pt}\setlength{\parskip}{0pt}
You are naming the object class of Wikimedia Commons categories. Each category is about one physical object (or one manufactured design). Your output is a single generic noun phrase per category -- the name of the kind of thing it is. It will later be used as a text prompt for an image segmentation model; you do not need to know anything about that model. Just name the object correctly and consistently.\par
\smallskip
Each numbered item gives the category TITLE, its parent category, the crawl root it was reached from, and a granularity label (individual / model / named\_unknown / pool) from an earlier pass.\par
\smallskip
For each item return:\par
\smallskip
sam\_prompt -- ONE short generic noun phrase for the object class: {\char34}statue{\char34}, {\char34}cannon{\char34}, {\char34}gravestone{\char34}, {\char34}pipe organ{\char34}, {\char34}telephone box{\char34}, {\char34}typewriter{\char34}, {\char34}wind turbine{\char34}. Rules:\par
\hangindent=1.2em\hangafter=1 - Generic, never specific. {\char34}Bust of Voltaire{\char34} -\textgreater{} {\char34}bust{\char34}. {\char34}Pentax K10D{\char34} -\textgreater{} {\char34}camera{\char34}. {\char34}K6 telephone boxes in Derbyshire{\char34} -\textgreater{} {\char34}telephone box{\char34}. Names, model numbers, places, and reigns never appear in the output.\par
\hangindent=1.2em\hangafter=1 - Name the WHOLE object the category is about, not a component: {\char34}pipe organ{\char34} not {\char34}organ pipes{\char34}; {\char34}fountain{\char34} not {\char34}basin{\char34}; {\char34}cannon{\char34} not {\char34}cannon barrel{\char34}.\par
\hangindent=1.2em\hangafter=1 - No material, color, pose, or style: {\char34}bronze equestrian statue{\char34} -\textgreater{} {\char34}statue{\char34}. Those describe one instance; the same word must fit every unit in the category.\par
\hangindent=1.2em\hangafter=1 - Two-word compounds are fine when a single word is wrong or too broad: {\char34}wayside cross{\char34}, {\char34}baptismal font{\char34}, {\char34}sewing machine{\char34}, {\char34}vending machine{\char34}.\par
\smallskip
ANCHOR TO THE CATEGORY TREE. The parent and root tell you the object{\char39}s family, and that is authoritative: everything under a {\char34}Dolmens{\char34} root is a dolmen, everything under {\char34}Pipe organs by country{\char34} is a pipe organ. Use the family{\char39}s noun unless the TITLE clearly names something more specific or different (a {\char34}War memorials{\char34} root can hold an obelisk, a statue, or a wall -- take the title{\char39}s word there). Root and parent names are index-shaped ({\char34}Coastal artillery{\char34}, {\char34}War memorials by country{\char34}) -- give the object noun ({\char34}cannon{\char34}, {\char34}war memorial{\char34}), not the index name. Be CONSISTENT: every category in the same family should get the same word unless its title says otherwise.\par
\smallskip
alternatives -- normally an EMPTY list. Fill it ONLY when the title and the tree genuinely disagree about what the object is, or the title alone does not make the object clear (e.g. a battery site that could be read as guns, a bunker, or an installation). A synonym of your chosen word is NOT an uncertainty ({\char34}dolmen{\char34} is not unsure because {\char34}megalithic tomb{\char34} exists) -- never list synonyms. At most 2 entries. A non-empty list marks the item for human review, so listing alternatives {\char34}to be safe{\char34} defeats its purpose.\par
\smallskip
confidence -- {\char34}high{\char34} when the object class is clear from title + tree, {\char34}low{\char34} when you filled alternatives or the title does not make the object clear. Nothing else.\par
\endgroup\par\nobreak\smallskip\noindent\rule{\columnwidth}{0.3pt}\par\medskip
\noindent User message:
\par\smallskip\noindent\rule{\columnwidth}{0.3pt}\par\nobreak\smallskip
\begingroup\scriptsize\ttfamily\RaggedRight\setlength{\parindent}{0pt}\setlength{\parskip}{0pt}
Assign segmentation prompts to these categories:\par
\smallskip
0. {\char34}\textless{}title\textgreater{}{\char34} | parent: \textless{}parent category\textgreater{} | root: \textless{}crawl root\textgreater{} | granularity: \textless{}label\textgreater{}\par
1. ...\par
\endgroup\par\nobreak\smallskip\noindent\rule{\columnwidth}{0.3pt}\par\medskip

\medskip\noindent\textbf{Root-category screening} (Section~\ref{sec:supp:curation}).
The response for each category gives a yes or no decision and a short reason.
\par\smallskip\noindent\rule{\columnwidth}{0.3pt}\par\nobreak\smallskip
\begingroup\scriptsize\ttfamily\RaggedRight\setlength{\parindent}{0pt}\setlength{\parskip}{0pt}
You are screening Wikimedia Commons category names to decide which ones are FAMILIES OF RIGID PHYSICAL OBJECTS worth crawling for a 3D object-reconstruction dataset. Each name is a direct subcategory of a broad {\char34}umbrella{\char34} category (e.g. {\char34}Musical instruments{\char34}, {\char34}Tools{\char34}, {\char34}Furniture{\char34}, {\char34}Containers{\char34}); the umbrella is given for context.\par
\smallskip
Answer YES when the category is a class of manufactured or carved rigid objects whose members would be photographed as objects: guitars, pianos, chairs, knives, watches, telephones, howitzers, loudspeakers, tins, buckets, radars, electric motors.\par
\smallskip
Answer NO for anything else:\par
\hangindent=1.2em\hangafter=1 - topics, activities, concepts, history ({\char34}Ball games{\char34}, {\char34}History of synthesizers{\char34}, {\char34}Human-computer interaction{\char34}, {\char34}Light art{\char34}, {\char34}Synth DIY{\char34})\par
\hangindent=1.2em\hangafter=1 - people and occupations ({\char34}Clockmakers{\char34}, {\char34}Sculptors{\char34})\par
\hangindent=1.2em\hangafter=1 - vehicles of any kind -- cars, trucks, ships, aircraft, trains, bicycles count as vehicles here ({\char34}Vehicles{\char34}, {\char34}Anchorhandling, tug and supply ships{\char34}, {\char34}Survey ships{\char34}) -- they are covered by a separate dataset\par
\hangindent=1.2em\hangafter=1 - soft / deformable / flat things (textiles, rugs, bags, clothing, flags, paper, coins, plaques)\par
\hangindent=1.2em\hangafter=1 - media about objects (photos-of, logos, {\char34}Numbers on racing vehicles{\char34}, stickers)\par
\hangindent=1.2em\hangafter=1 - categories that are groupings by attribute rather than a family ({\char34}Metal sundials{\char34}, {\char34}Black liftbacks{\char34}, {\char34}...by color{\char34}, {\char34}...by year{\char34})\par
\hangindent=1.2em\hangafter=1 - ammunition cartridges are borderline small objects -- answer YES for guns/weapons families, NO for cartridges/bullets\par
\smallskip
When the name is a sub-family of something obviously object-like ({\char34}Bullpup{\char34}, {\char34}Guitars by type{\char34}, {\char34}Submachine guns by model{\char34}, {\char34}Lenses by brand{\char34}, {\char34}Watches by brand{\char34}) answer YES -- index suffixes ({\char34}by type/brand/model{\char34}) are fine; those are how Commons lists the members.\par
\smallskip
Output for each item: index, decision ({\char34}yes{\char34}/{\char34}no{\char34}), and a reason of at most 6 words.\par
\endgroup\par\nobreak\smallskip\noindent\rule{\columnwidth}{0.3pt}\par\medskip
\noindent User message:
\par\smallskip\noindent\rule{\columnwidth}{0.3pt}\par\nobreak\smallskip
\begingroup\scriptsize\ttfamily\RaggedRight\setlength{\parindent}{0pt}\setlength{\parskip}{0pt}
Screen these categories:\par
\smallskip
0. {\char34}\textless{}title\textgreater{}{\char34} | umbrella: \textless{}umbrella category\textgreater{}\par
1. ...\par
\endgroup\par\nobreak\smallskip\noindent\rule{\columnwidth}{0.3pt}\par\medskip

%% file: sec/S8_floats.tex
\begin{figure*}[p]
\centering\includegraphics[width=\textwidth]{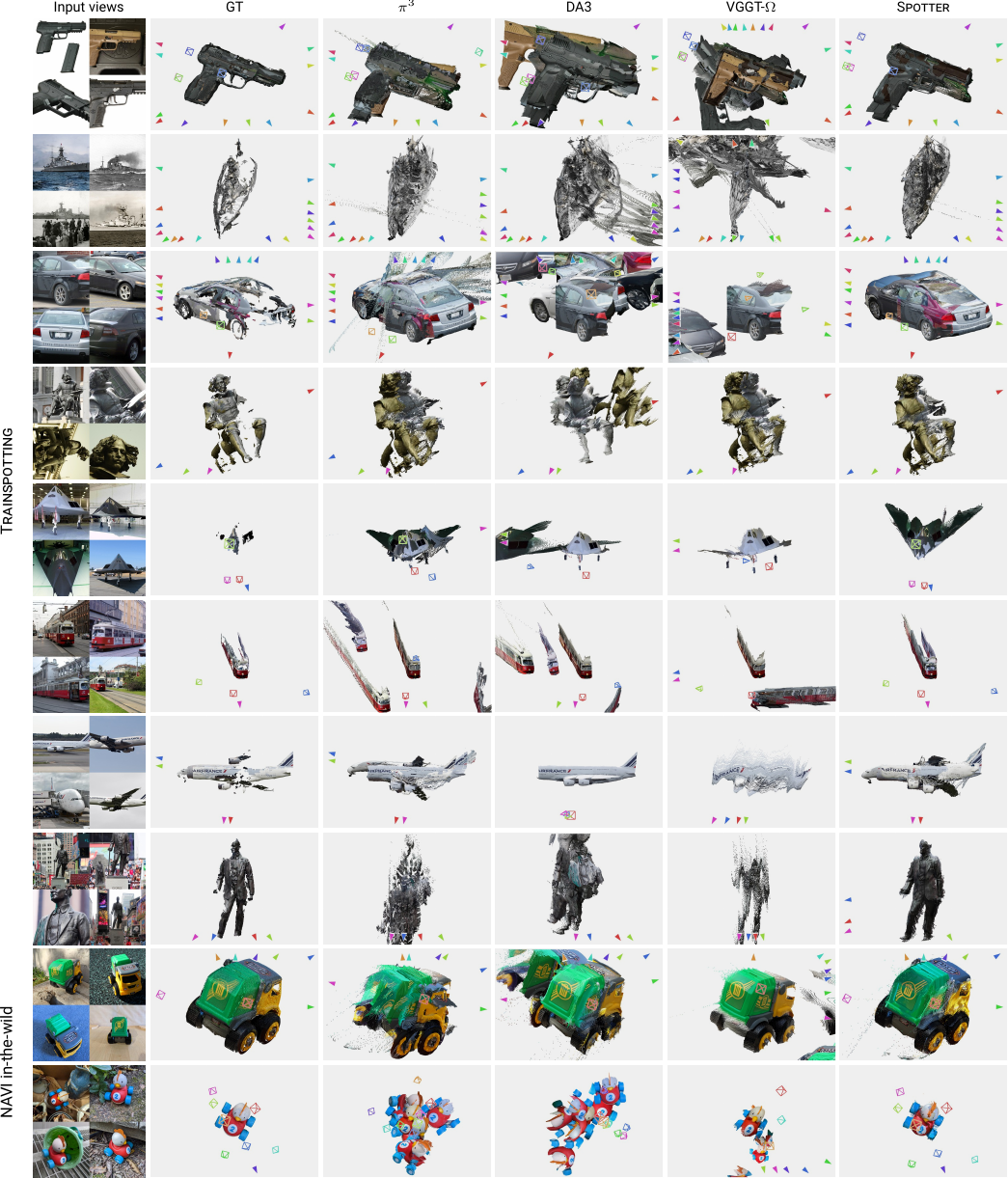}
\caption{\textbf{Additional qualitative comparisons.}
Ten held-out samples drawn at random: three \datasetname{} objects with 16 input views, five with 4 input views, and two NAVI in-the-wild objects with 8 input views.}
\label{fig:supp-random}
\end{figure*}
\input{tables/supp_object_full}
\input{tables/supp_external_full}

%% file: tables/supp_object_full.tex
\begin{table*}[p]
\caption{\textbf{All metrics on \datasetname{}.} Extends Table~3 of the main paper with the median of accuracy and completeness, the inlier ratios at 1.05 and 1.10, and AUC at $15^\circ$. Accuracy and completeness are in percent of the object extent.}
\label{tab:supp-object-full}
\centering\scriptsize\setlength{\tabcolsep}{3pt}
\begin{tabular*}{\textwidth}{@{\extracolsep{\fill}}l*{11}{c}@{}}
\toprule
Method & \multicolumn{4}{c}{Point map} & \multicolumn{4}{c}{Depth} & \multicolumn{3}{c}{Camera} \\
\cmidrule(lr){2-5} \cmidrule(lr){6-9} \cmidrule(lr){10-12}
 & Acc mean & Acc med. & Comp mean & Comp med. & AbsRel & $\delta_{1.03}$ & $\delta_{1.05}$ & $\delta_{1.10}$ & AUC$_5$ & AUC$_{15}$ & AUC$_{30}$ \\
\midrule
VGGT~\citep{wang2025vggt} & 2.38 & 1.62 & 8.08 & 6.48 & 0.160 & 32.4 & 41.8 & 56.2 & 11.5 & 29.9 & 44.1 \\
$\pi^3$~\citep{wang2026pi3} & 2.33 & 1.61 & 6.22 & 4.76 & 0.162 & 34.7 & 43.8 & 56.8 & 12.5 & 31.9 & 46.4 \\
DA3~\citep{lin2026da3} & 2.22 & 1.60 & 12.17 & 10.11 & 0.189 & 32.1 & 41.5 & 54.6 & 10.1 & 21.0 & 29.4 \\
MapAny.~\citep{keetha2026mapanything} & 2.20 & 1.57 & 7.75 & 6.18 & 0.206 & 26.6 & 35.3 & 49.1 & 5.3 & 19.3 & 33.1 \\
VGGT-$\Omega$~\citep{wang2026vggtomega} & 1.84 & 1.36 & 12.68 & 10.77 & 0.201 & 28.1 & 37.1 & 50.3 & 7.2 & 16.9 & 25.0 \\
\modelname{} & \textbf{1.22} & \textbf{0.89} & \textbf{1.15} & \textbf{0.81} & \textbf{0.071} & \textbf{53.3} & \textbf{65.4} & \textbf{80.5} & \textbf{44.2} & \textbf{70.1} & \textbf{81.2} \\
\bottomrule
\end{tabular*}
\vspace{-1em}
\end{table*}

%% file: tables/supp_external_full.tex
\begin{table*}[p]
\caption{\textbf{All metrics on the external benchmarks.} The metrics of Table~\ref{tab:supp-object-full} and normal consistency (NC) on the benchmarks of Tables~5 and~6 of the main paper. Accuracy and completeness are in millimeters.}
\label{tab:supp-external-full}
\centering\scriptsize\setlength{\tabcolsep}{3pt}\renewcommand{\arraystretch}{0.9}
\begin{tabular*}{\textwidth}{@{\extracolsep{\fill}}l*{12}{c}@{}}
\toprule
Method & \multicolumn{5}{c}{Point map} & \multicolumn{4}{c}{Depth} & \multicolumn{3}{c}{Camera} \\ \cmidrule(lr){2-6} \cmidrule(lr){7-10} \cmidrule(lr){11-13}
 & Acc mean & Acc med. & Comp mean & Comp med. & NC & AbsRel & $\delta_{1.03}$ & $\delta_{1.05}$ & $\delta_{1.10}$ & AUC$_5$ & AUC$_{15}$ & AUC$_{30}$ \\
\midrule
\multicolumn{13}{l}{\textit{7-Scenes (18 samples)}} \\
VGGT~\citep{wang2025vggt} & 25.3 & 11.6 & 32.6 & 15.9 & 0.735 & \textbf{0.277} & 67.5 & 83.4 & 92.3 & 26.6 & 65.2 & 81.0 \\
$\pi^3$~\citep{wang2026pi3} & \textbf{20.0} & 10.5 & \textbf{26.7} & 14.3 & \textbf{0.765} & 0.342 & 70.5 & 87.1 & \textbf{96.4} & 28.8 & 68.5 & 83.3 \\
DA3~\citep{lin2026da3} & 25.4 & 13.1 & 35.7 & 15.9 & 0.755 & 0.324 & 74.1 & 88.1 & 94.6 & 28.5 & 66.3 & 81.5 \\
MapAny.~\citep{keetha2026mapanything} & 28.2 & 14.7 & 36.8 & 16.0 & 0.739 & 0.332 & 58.3 & 77.3 & 93.3 & 18.2 & 54.7 & 72.4 \\
VGGT-$\Omega$~\citep{wang2026vggtomega} & 21.8 & 11.9 & 27.4 & 14.4 & 0.760 & 0.295 & 71.9 & 86.9 & 94.3 & \textbf{40.8} & \textbf{75.4} & \textbf{87.2} \\
\modelname{} & 20.2 & \textbf{10.2} & 27.4 & \textbf{14.1} & 0.760 & 0.301 & \textbf{75.1} & \textbf{88.3} & 94.5 & 38.8 & 73.8 & 86.1 \\
\midrule
\multicolumn{13}{l}{\textit{NRGBD (9 samples)}} \\
VGGT~\citep{wang2025vggt} & 17.4 & 10.1 & 16.9 & 7.0 & 0.914 & 0.016 & 90.9 & 96.2 & 98.3 & 86.4 & 95.1 & 97.6 \\
$\pi^3$~\citep{wang2026pi3} & 17.9 & 9.3 & 17.2 & 6.8 & 0.913 & 0.016 & 88.6 & 94.6 & 98.3 & 86.5 & 94.7 & 97.2 \\
DA3~\citep{lin2026da3} & 16.4 & 7.8 & 15.3 & 6.3 & \textbf{0.932} & \textbf{0.010} & \textbf{96.7} & \textbf{98.0} & 98.6 & \textbf{91.4} & \textbf{97.1} & \textbf{98.6} \\
MapAny.~\citep{keetha2026mapanything} & 42.8 & 22.2 & 30.2 & 12.5 & 0.856 & 0.033 & 69.2 & 85.9 & 95.5 & 56.5 & 82.4 & 90.9 \\
VGGT-$\Omega$~\citep{wang2026vggtomega} & 13.3 & 8.2 & 9.0 & 5.7 & 0.908 & 0.013 & 92.4 & \textbf{98.0} & \textbf{99.1} & 87.8 & 95.7 & 97.9 \\
\modelname{} & \textbf{12.9} & \textbf{7.7} & \textbf{8.3} & \textbf{5.5} & 0.914 & 0.013 & 92.7 & 97.7 & 99.0 & 86.5 & 95.1 & 97.3 \\
\midrule
\multicolumn{13}{l}{\textit{DTU (22 samples)}} \\
VGGT~\citep{wang2025vggt} & 1.2 & 0.7 & 1.8 & 1.0 & 0.699 & \textbf{0.009} & \textbf{97.4} & \textbf{98.0} & 98.6 & \textbf{98.1} & \textbf{99.4} & \textbf{99.7} \\
$\pi^3$~\citep{wang2026pi3} & \textbf{1.2} & 0.7 & \textbf{1.5} & \textbf{0.6} & 0.695 & 0.014 & 91.2 & 96.9 & \textbf{98.7} & 80.0 & 93.2 & 96.6 \\
DA3~\citep{lin2026da3} & 1.7 & 1.0 & 1.6 & 0.8 & \textbf{0.711} & 0.009 & 96.5 & 97.9 & 98.6 & 91.0 & 97.0 & 98.5 \\
MapAny.~\citep{keetha2026mapanything} & 2.4 & 1.3 & 1.9 & 0.8 & 0.702 & 0.067 & 37.8 & 55.2 & 77.8 & 15.0 & 53.3 & 73.4 \\
VGGT-$\Omega$~\citep{wang2026vggtomega} & 1.2 & \textbf{0.7} & 1.7 & 0.6 & 0.701 & 0.009 & 97.1 & \textbf{98.0} & \textbf{98.7} & 91.1 & 97.0 & 98.5 \\
\modelname{} & 1.2 & 0.7 & 1.6 & 0.6 & 0.700 & 0.012 & 95.9 & 97.5 & 98.6 & 86.1 & 95.3 & 97.7 \\
\midrule
\multicolumn{13}{l}{\textit{ETH3D (13 samples)}} \\
VGGT~\citep{wang2025vggt} & 291.4 & 190.7 & 317.5 & 204.4 & 0.841 & 0.045 & 52.6 & 69.5 & 88.4 & 37.6 & 58.5 & 71.1 \\
$\pi^3$~\citep{wang2026pi3} & 154.7 & 87.2 & 129.7 & 65.1 & 0.890 & 0.032 & 62.4 & 78.4 & 93.4 & 36.4 & 58.7 & 72.9 \\
DA3~\citep{lin2026da3} & 274.3 & 197.3 & 270.3 & 170.1 & 0.844 & 0.027 & 73.2 & 84.8 & 95.4 & 55.4 & 69.5 & 75.4 \\
MapAny.~\citep{keetha2026mapanything} & 459.7 & 377.7 & 542.0 & 413.9 & 0.812 & 0.161 & 51.9 & 65.7 & 81.3 & 33.8 & 54.4 & 65.6 \\
VGGT-$\Omega$~\citep{wang2026vggtomega} & 73.6 & 43.5 & 61.1 & 34.5 & 0.915 & \textbf{0.017} & \textbf{83.5} & \textbf{94.5} & \textbf{99.2} & \textbf{66.9} & \textbf{84.9} & \textbf{91.8} \\
\modelname{} & \textbf{65.5} & \textbf{37.1} & \textbf{52.9} & \textbf{30.3} & \textbf{0.924} & 0.017 & 82.8 & 94.3 & 98.7 & 65.4 & 84.8 & \textbf{91.8} \\
\midrule
\multicolumn{13}{l}{\textit{NAVI (281 samples)}} \\
VGGT~\citep{wang2025vggt} & 2.5 & 1.6 & 3.4 & 2.7 & 0.751 & 0.032 & 71.5 & 87.1 & 96.3 & 53.3 & 82.7 & 91.2 \\
$\pi^3$~\citep{wang2026pi3} & \textbf{2.2} & \textbf{1.4} & \textbf{3.0} & \textbf{2.4} & \textbf{0.762} & \textbf{0.027} & 76.5 & 91.3 & \textbf{97.5} & 55.8 & 83.6 & 91.6 \\
DA3~\citep{lin2026da3} & 3.3 & 2.0 & 5.7 & 4.2 & 0.682 & 0.034 & \textbf{80.0} & \textbf{91.9} & 97.1 & \textbf{57.2} & 84.1 & 91.8 \\
MapAny.~\citep{keetha2026mapanything} & 3.8 & 2.3 & 5.8 & 4.6 & 0.663 & 0.051 & 56.4 & 75.3 & 91.9 & 28.4 & 69.1 & 84.0 \\
VGGT-$\Omega$~\citep{wang2026vggtomega} & 2.9 & 1.8 & 5.3 & 4.1 & 0.692 & 0.034 & 76.1 & 90.5 & 97.1 & 55.4 & 83.6 & 91.6 \\
\modelname{} & 3.1 & 1.9 & 4.0 & 3.0 & 0.702 & 0.032 & 74.5 & 89.9 & 97.2 & 56.7 & \textbf{84.2} & \textbf{91.9} \\
\midrule
\multicolumn{13}{l}{\textit{NAVI (wild) (35 samples)}} \\
VGGT~\citep{wang2025vggt} & 3.2 & 2.4 & 11.5 & 10.0 & 0.633 & 0.085 & 35.0 & 48.1 & 70.8 & 14.4 & 34.8 & 45.5 \\
$\pi^3$~\citep{wang2026pi3} & 3.7 & 2.8 & 9.6 & 8.1 & 0.628 & 0.078 & 41.2 & 55.3 & 77.2 & 13.1 & 32.1 & 43.7 \\
DA3~\citep{lin2026da3} & 3.3 & 2.5 & 16.0 & 14.0 & 0.578 & 0.079 & 41.1 & 56.9 & 76.7 & 12.0 & 28.5 & 36.9 \\
MapAny.~\citep{keetha2026mapanything} & 4.1 & 2.9 & 11.0 & 9.5 & 0.588 & 0.103 & 29.6 & 43.4 & 66.9 & 3.3 & 14.7 & 24.2 \\
VGGT-$\Omega$~\citep{wang2026vggtomega} & \textbf{3.1} & 2.5 & 20.4 & 18.5 & 0.544 & 0.101 & 29.7 & 44.6 & 66.9 & 3.8 & 11.6 & 18.8 \\
\modelname{} & 3.2 & \textbf{2.1} & \textbf{1.9} & \textbf{1.2} & \textbf{0.743} & \textbf{0.056} & \textbf{43.1} & \textbf{60.9} & \textbf{82.6} & \textbf{30.7} & \textbf{62.2} & \textbf{76.0} \\
\bottomrule
\end{tabular*}
\vspace{-1em}
\end{table*}

%% file: arxiv.bbl
\begin{thebibliography}{65}
\providecommand{\natexlab}[1]{#1}
\providecommand{\url}[1]{\texttt{#1}}
\expandafter\ifx\csname urlstyle\endcsname\relax
  \providecommand{\doi}[1]{doi: #1}\else
  \providecommand{\doi}{doi: \begingroup \urlstyle{rm}\Url}\fi

\bibitem[Aan{\ae}s et~al.(2016)Aan{\ae}s, Jensen, Vogiatzis, Tola, and
  Dahl]{aanaes2016dtu}
Henrik Aan{\ae}s, Rasmus~Ramsb{\o}l Jensen, George Vogiatzis, Engin Tola, and
  Anders~Bjorholm Dahl.
\newblock Large-scale data for multiple-view stereopsis.
\newblock \emph{IJCV}, 120:\penalty0 153--168, 2016.

\bibitem[Agarwal et~al.(2009)Agarwal, Snavely, Simon, Seitz, and
  Szeliski]{agarwal2009rome}
Sameer Agarwal, Noah Snavely, Ian Simon, Steven~M. Seitz, and Richard Szeliski.
\newblock Building rome in a day.
\newblock In \emph{ICCV}, 2009.

\bibitem[Arandjelovi\'{c} et~al.(2016)Arandjelovi\'{c}, Gronat, Torii, Pajdla,
  and Sivic]{arandjelovic2016netvlad}
Relja Arandjelovi\'{c}, Petr Gronat, Akihiko Torii, Tomas Pajdla, and Josef
  Sivic.
\newblock {NetVLAD}: {CNN} architecture for weakly supervised place
  recognition.
\newblock In \emph{CVPR}, 2016.

\bibitem[Azinovi{\'c} et~al.(2022)Azinovi{\'c}, Martin-Brualla, Goldman,
  Nie{\ss}ner, and Thies]{azinovic2022nrgbd}
Dejan Azinovi{\'c}, Ricardo Martin-Brualla, Dan~B Goldman, Matthias
  Nie{\ss}ner, and Justus Thies.
\newblock Neural {RGB-D} surface reconstruction.
\newblock In \emph{CVPR}, 2022.

\bibitem[Barath et~al.(2020)Barath, Noskova, Ivashechkin, and
  Matas]{barath2020magsacpp}
Daniel Barath, Jana Noskova, Maksym Ivashechkin, and Ji{\v r}{\'\i} Matas.
\newblock {MAGSAC++}, a fast, reliable and accurate robust estimator.
\newblock In \emph{CVPR}, 2020.

\bibitem[Carion et~al.(2026)Carion, Gustafson, Hu, Debnath, Hu, Suris, Ryali,
  Alwala, Khedr, Huang, Lei, Ma, Guo, Kalla, Marks, Greer, Wang, Sun,
  R{\"a}dle, Afouras, Mavroudi, Xu, Wu, Zhou, Momeni, Hazra, Ding, Vaze,
  Porcher, Li, Li, Kamath, Cheng, Doll{\'a}r, Ravi, Saenko, Zhang, and
  Feichtenhofer]{sam3}
Nicolas Carion, Laura Gustafson, Yuan-Ting Hu, Shoubhik Debnath, Ronghang Hu,
  Didac Suris, Chaitanya Ryali, Kalyan~Vasudev Alwala, Haitham Khedr, Andrew
  Huang, Jie Lei, Tengyu Ma, Baishan Guo, Arpit Kalla, Markus Marks, Joseph
  Greer, Meng Wang, Peize Sun, Roman R{\"a}dle, Triantafyllos Afouras,
  Effrosyni Mavroudi, Katherine Xu, Tsung-Han Wu, Yu Zhou, Liliane Momeni,
  Rishi Hazra, Shuangrui Ding, Sagar Vaze, Francois Porcher, Feng Li, Siyuan
  Li, Aishwarya Kamath, Ho~Kei Cheng, Piotr Doll{\'a}r, Nikhila Ravi, Kate
  Saenko, Pengchuan Zhang, and Christoph Feichtenhofer.
\newblock {SAM} 3: Segment anything with concepts.
\newblock In \emph{ICLR}, 2026.

\bibitem[Chang et~al.(2015)Chang, Funkhouser, Guibas, Hanrahan, Huang, Li,
  Savarese, Savva, Song, Su, Xiao, Yi, and Yu]{chang2015shapenet}
Angel~X. Chang, Thomas Funkhouser, Leonidas Guibas, Pat Hanrahan, Qixing Huang,
  Zimo Li, Silvio Savarese, Manolis Savva, Shuran Song, Hao Su, Jianxiong Xiao,
  Li Yi, and Fisher Yu.
\newblock {ShapeNet}: An information-rich {3D} model repository.
\newblock \emph{arXiv preprint arXiv:1512.03012}, 2015.

\bibitem[Dai et~al.(2017)Dai, Chang, Savva, Halber, Funkhouser, and
  Nie{\ss}ner]{dai2017scannet}
Angela Dai, Angel~X. Chang, Manolis Savva, Maciej Halber, Thomas Funkhouser,
  and Matthias Nie{\ss}ner.
\newblock Scannet: Richly-annotated 3d reconstructions of indoor scenes.
\newblock In \emph{CVPR}, 2017.

\bibitem[Deitke et~al.(2023{\natexlab{a}})Deitke, Liu, Wallingford, Ngo,
  Michel, Kusupati, Fan, Laforte, Voleti, Gadre, VanderBilt, Kembhavi,
  Vondrick, Gkioxari, Ehsani, Schmidt, and Farhadi]{deitke2023objaversexl}
Matt Deitke, Ruoshi Liu, Matthew Wallingford, Huong Ngo, Oscar Michel, Aditya
  Kusupati, Alan Fan, Christian Laforte, Vikram Voleti, Samir~Yitzhak Gadre,
  Eli VanderBilt, Aniruddha Kembhavi, Carl Vondrick, Georgia Gkioxari, Kiana
  Ehsani, Ludwig Schmidt, and Ali Farhadi.
\newblock {Objaverse-XL}: A universe of 10m+ {3D} objects.
\newblock In \emph{NeurIPS Datasets and Benchmarks}, 2023{\natexlab{a}}.

\bibitem[Deitke et~al.(2023{\natexlab{b}})Deitke, Schwenk, Salvador, Weihs,
  Michel, VanderBilt, Schmidt, Ehsani, Kembhavi, and
  Farhadi]{deitke2023objaverse}
Matt Deitke, Dustin Schwenk, Jordi Salvador, Luca Weihs, Oscar Michel, Eli
  VanderBilt, Ludwig Schmidt, Kiana Ehsani, Aniruddha Kembhavi, and Ali
  Farhadi.
\newblock Objaverse: A universe of annotated 3d objects.
\newblock In \emph{CVPR}, 2023{\natexlab{b}}.

\bibitem[DeTone et~al.(2018)DeTone, Malisiewicz, and
  Rabinovich]{detone2018superpoint}
Daniel DeTone, Tomasz Malisiewicz, and Andrew Rabinovich.
\newblock Superpoint: Self-supervised interest point detection and description.
\newblock In \emph{CVPR Workshops}, 2018.

\bibitem[Downs et~al.(2022)Downs, Francis, Koenig, Kinman, Hickman, Reymann,
  McHugh, and Vanhoucke]{downs2022gso}
Laura Downs, Anthony Francis, Nate Koenig, Brandon Kinman, Ryan Hickman, Krista
  Reymann, Thomas~B. McHugh, and Vincent Vanhoucke.
\newblock {Google Scanned Objects}: A high-quality dataset of {3D} scanned
  household items.
\newblock In \emph{ICRA}, 2022.

\bibitem[Duisterhof et~al.(2025)Duisterhof, Zust, Weinzaepfel, Leroy, Cabon,
  and Revaud]{duisterhof2024mast3rsfm}
Bardienus Duisterhof, Lojze Zust, Philippe Weinzaepfel, Vincent Leroy, Yohann
  Cabon, and J{\'e}r{\^o}me Revaud.
\newblock {MASt3R-SfM}: a fully-integrated solution for unconstrained
  structure-from-motion.
\newblock In \emph{3DV}, 2025.

\bibitem[Edstedt et~al.(2023)Edstedt, Athanasiadis, Wadenb{\"a}ck, and
  Felsberg]{edstedt2023dkm}
Johan Edstedt, Ioannis Athanasiadis, M{\r{a}}rten Wadenb{\"a}ck, and Michael
  Felsberg.
\newblock Dkm: Dense kernelized feature matching for geometry estimation.
\newblock In \emph{CVPR}, 2023.

\bibitem[Edstedt et~al.(2024)Edstedt, Sun, B{\"o}kman, Wadenb{\"a}ck, and
  Felsberg]{edstedt2024roma}
Johan Edstedt, Qiyu Sun, Georg B{\"o}kman, M{\r{a}}rten Wadenb{\"a}ck, and
  Michael Felsberg.
\newblock Roma: Robust dense feature matching.
\newblock In \emph{CVPR}, 2024.

\bibitem[Edstedt et~al.(2025)Edstedt, Nordstr{\"o}m, Zhang, B{\"o}kman,
  Astermark, Larsson, Heyden, Kahl, Wadenb{\"a}ck, and Felsberg]{romav2}
Johan Edstedt, David Nordstr{\"o}m, Yushan Zhang, Georg B{\"o}kman, Jonathan
  Astermark, Viktor Larsson, Anders Heyden, Fredrik Kahl, M{\r{a}}rten
  Wadenb{\"a}ck, and Michael Felsberg.
\newblock {RoMa v2}: Harder better faster denser feature matching, 2025.
\newblock arXiv:2511.15706.

\bibitem[He et~al.(2022)He, Sun, Wang, Huang, Bao, and Zhou]{he2022oneposepp}
Xingyi He, Jiaming Sun, Yuang Wang, Di Huang, Hujun Bao, and Xiaowei Zhou.
\newblock {OnePose++}: Keypoint-free one-shot object pose estimation without
  {CAD} models.
\newblock In \emph{NeurIPS}, 2022.

\bibitem[He et~al.(2024)He, Sun, Wang, Peng, Huang, Bao, and Zhou]{he2024dfsfm}
Xingyi He, Jiaming Sun, Yifan Wang, Sida Peng, Qixing Huang, Hujun Bao, and
  Xiaowei Zhou.
\newblock Detector-free structure from motion.
\newblock In \emph{CVPR}, 2024.

\bibitem[Jampani et~al.(2023)Jampani, Maninis, Engelhardt, Karpur, Truong,
  Sargent, Popov, Araujo, Martin-Brualla, Patel, Vlasic, Ferrari, Makadia, Liu,
  Li, and Zhou]{jampani2023navi}
Varun Jampani, Kevis-Kokitsi Maninis, Andreas Engelhardt, Arjun Karpur, Karen
  Truong, Kyle Sargent, Stefan Popov, Andr{\'e} Araujo, Ricardo Martin-Brualla,
  Kaushal Patel, Daniel Vlasic, Vittorio Ferrari, Ameesh Makadia, Ce Liu,
  Yuanzhen Li, and Howard Zhou.
\newblock {NAVI}: Category-agnostic image collections with high-quality {3D}
  shape and pose annotations.
\newblock In \emph{NeurIPS Datasets and Benchmarks}, 2023.

\bibitem[Keetha et~al.(2026)Keetha, M{\"u}ller, Sch{\"o}nberger, Porzi, Zhang,
  Fischer, Knapitsch, Zauss, Weber, Antunes, Luiten, Lopez-Antequera,
  Rota~Bul{\`o}, Richardt, Ramanan, Scherer, and
  Kontschieder]{keetha2026mapanything}
Nikhil Keetha, Norman M{\"u}ller, Johannes Sch{\"o}nberger, Lorenzo Porzi,
  Yuchen Zhang, Tobias Fischer, Arno Knapitsch, Duncan Zauss, Ethan Weber,
  Nelson Antunes, Jonathon Luiten, Manuel Lopez-Antequera, Samuel
  Rota~Bul{\`o}, Christian Richardt, Deva Ramanan, Sebastian Scherer, and Peter
  Kontschieder.
\newblock {MapAnything}: Universal feed-forward metric {3D} reconstruction.
\newblock In \emph{3DV}, 2026.

\bibitem[Lee and Yoo(2025)]{lee2025densesfm}
JongMin Lee and Sungjoo Yoo.
\newblock {Dense-SfM}: Structure from motion with dense consistent matching.
\newblock In \emph{CVPR}, 2025.

\bibitem[Leroy et~al.(2024)Leroy, Cabon, and Revaud]{leroy2024mast3r}
Vincent Leroy, Yohann Cabon, and J{\'e}r{\^o}me Revaud.
\newblock Grounding image matching in 3d with {MASt3R}.
\newblock In \emph{ECCV}, 2024.

\bibitem[Li et~al.(2026)Li, Xiangli, Averbuch-Elor, Snavely, and
  Cai]{li2026megadepthx}
Yuan Li, Yuanbo Xiangli, Hadar Averbuch-Elor, Noah Snavely, and Ruojin Cai.
\newblock Long-tail internet photo reconstruction.
\newblock In \emph{CVPR}, 2026.

\bibitem[Li and Snavely(2018)]{li2018megadepth}
Zhengqi Li and Noah Snavely.
\newblock Megadepth: Learning single-view depth prediction from internet
  photos.
\newblock In \emph{CVPR}, 2018.

\bibitem[Lin et~al.(2026)Lin, Chen, Liew, Chen, Li, Zhao, Peng, Guo, Zhou, Shi,
  Feng, and Kang]{lin2026da3}
Haotong Lin, Sili Chen, Jun~Hao Liew, Donny~Y. Chen, Zhenyu Li, Yang Zhao, Sida
  Peng, Hengkai Guo, Xiaowei Zhou, Guang Shi, Jiashi Feng, and Bingyi Kang.
\newblock Depth anything 3: Recovering the visual space from any views.
\newblock In \emph{ICLR}, 2026.

\bibitem[Lindenberger et~al.(2023)Lindenberger, Sarlin, and
  Pollefeys]{lindenberger2023lightglue}
Philipp Lindenberger, Paul-Edouard Sarlin, and Marc Pollefeys.
\newblock Lightglue: Local feature matching at light speed.
\newblock In \emph{ICCV}, 2023.

\bibitem[Ling et~al.(2024)Ling, Sheng, Tu, Zhao, Xin, Wan, Yu, Guo, Yu, Lu, Li,
  Sun, Ashok, Mukherjee, Kang, Kong, Hua, Zhang, Benes, and
  Bera]{ling2024dl3dv}
Lu Ling, Yichen Sheng, Zhi Tu, Wentian Zhao, Cheng Xin, Kun Wan, Lantao Yu,
  Qianyu Guo, Zixun Yu, Yawen Lu, Xuanmao Li, Xingpeng Sun, Rohan Ashok,
  Aniruddha Mukherjee, Hao Kang, Xiangrui Kong, Gang Hua, Tianyi Zhang, Bedrich
  Benes, and Aniket Bera.
\newblock {DL3DV-10K}: A large-scale scene dataset for deep learning-based {3D}
  vision.
\newblock In \emph{CVPR}, 2024.

\bibitem[Liu et~al.(2025)Liu, Tayal, Wang, Zarzar, Monnier, Tertikas, Duan,
  Toisoul, Zhang, Neverova, Vedaldi, Shapovalov, and Novotny]{liu2025uco3d}
Xingchen Liu, Piyush Tayal, Jianyuan Wang, Jesus Zarzar, Tom Monnier,
  Konstantinos Tertikas, Jiali Duan, Antoine Toisoul, Jason~Y. Zhang, Natalia
  Neverova, Andrea Vedaldi, Roman Shapovalov, and David Novotny.
\newblock {UnCommon} objects in {3D}.
\newblock In \emph{CVPR}, 2025.

\bibitem[Loshchilov and Hutter(2019)]{loshchilov2019adamw}
Ilya Loshchilov and Frank Hutter.
\newblock Decoupled weight decay regularization.
\newblock In \emph{ICLR}, 2019.

\bibitem[Lowe(2004)]{lowe2004sift}
David~G. Lowe.
\newblock Distinctive image features from scale-invariant keypoints.
\newblock \emph{IJCV}, 60\penalty0 (2):\penalty0 91--110, 2004.

\bibitem[Reizenstein et~al.(2021)Reizenstein, Shapovalov, Henzler, Sbordone,
  Labatut, and Novotny]{reizenstein2021co3d}
Jeremy Reizenstein, Roman Shapovalov, Philipp Henzler, Luca Sbordone, Patrick
  Labatut, and David Novotny.
\newblock Common objects in 3d: Large-scale learning and evaluation of
  real-life 3d category reconstruction.
\newblock In \emph{ICCV}, 2021.

\bibitem[Roberts et~al.(2021)Roberts, Ramapuram, Ranjan, Kumar, Bautista,
  Paczan, Webb, and Susskind]{roberts2021hypersim}
Mike Roberts, Jason Ramapuram, Anurag Ranjan, Atulit Kumar, Miguel~Angel
  Bautista, Nathan Paczan, Russ Webb, and Joshua~M Susskind.
\newblock {Hypersim}: A photorealistic synthetic dataset for holistic indoor
  scene understanding.
\newblock In \emph{ICCV}, 2021.

\bibitem[Sarlin et~al.(2019)Sarlin, Cadena, Siegwart, and
  Dymczyk]{sarlin2019coarse}
Paul-Edouard Sarlin, Cesar Cadena, Roland Siegwart, and Marcin Dymczyk.
\newblock From coarse to fine: Robust hierarchical localization at large scale.
\newblock In \emph{CVPR}, 2019.

\bibitem[Sarlin et~al.(2020)Sarlin, DeTone, Malisiewicz, and
  Rabinovich]{sarlin2020superglue}
Paul-Edouard Sarlin, Daniel DeTone, Tomasz Malisiewicz, and Andrew Rabinovich.
\newblock Superglue: Learning feature matching with graph neural networks.
\newblock In \emph{CVPR}, 2020.

\bibitem[Sch{\"o}nberger and Frahm(2016)]{schoenberger2016sfm}
Johannes~L. Sch{\"o}nberger and Jan-Michael Frahm.
\newblock Structure-from-motion revisited.
\newblock In \emph{CVPR}, 2016.

\bibitem[Sch{\"o}ps et~al.(2017)Sch{\"o}ps, Sch{\"o}nberger, Galliani, Sattler,
  Schindler, Pollefeys, and Geiger]{schops2017eth3d}
Thomas Sch{\"o}ps, Johannes~L Sch{\"o}nberger, Silvano Galliani, Torsten
  Sattler, Konrad Schindler, Marc Pollefeys, and Andreas Geiger.
\newblock A multi-view stereo benchmark with high-resolution images and
  multi-camera videos.
\newblock In \emph{CVPR}, 2017.

\bibitem[Shotton et~al.(2013)Shotton, Glocker, Zach, Izadi, Criminisi, and
  Fitzgibbon]{shotton2013sevenscenes}
Jamie Shotton, Ben Glocker, Christopher Zach, Shahram Izadi, Antonio Criminisi,
  and Andrew Fitzgibbon.
\newblock Scene coordinate regression forests for camera relocalization in
  {RGB-D} images.
\newblock In \emph{CVPR}, 2013.

\bibitem[Snavely et~al.(2006)Snavely, Seitz, and
  Szeliski]{snavely2006phototourism}
Noah Snavely, Steven~M. Seitz, and Richard Szeliski.
\newblock Photo tourism: Exploring photo collections in {3D}.
\newblock \emph{ACM Transactions on Graphics}, 25\penalty0 (3):\penalty0
  835--846, 2006.

\bibitem[Sun et~al.(2021)Sun, Shen, Wang, Bao, and Zhou]{sun2021loftr}
Jiaming Sun, Zehong Shen, Yuang Wang, Hujun Bao, and Xiaowei Zhou.
\newblock {LoFTR}: Detector-free local feature matching with transformers.
\newblock In \emph{CVPR}, 2021.

\bibitem[Sun et~al.(2018)Sun, Wu, Zhang, Zhang, Zhang, Xue, Tenenbaum, and
  Freeman]{sun2018pix3d}
Xingyuan Sun, Jiajun Wu, Xiuming Zhang, Zhoutong Zhang, Chengkai Zhang, Tianfan
  Xue, Joshua~B. Tenenbaum, and William~T. Freeman.
\newblock {Pix3D}: Dataset and methods for single-image {3D} shape modeling.
\newblock In \emph{CVPR}, 2018.

\bibitem[Tang et~al.(2025)Tang, Fan, Wang, Xu, Ranjan, Schwing, and
  Yan]{tang2025mvdust3r}
Zhenggang Tang, Yuchen Fan, Dilin Wang, Hongyu Xu, Rakesh Ranjan, Alexander
  Schwing, and Zhicheng Yan.
\newblock {MV-DUSt3R+}: Single-stage scene reconstruction from sparse views in
  2 seconds.
\newblock In \emph{CVPR}, 2025.

\bibitem[Tung et~al.(2024)Tung, Chou, Cai, Yang, Zhang, Wetzstein, Hariharan,
  and Snavely]{tung2024megascenes}
Joseph Tung, Gene Chou, Ruojin Cai, Guandao Yang, Kai Zhang, Gordon Wetzstein,
  Bharath Hariharan, and Noah Snavely.
\newblock Megascenes: Scene-level view synthesis at scale.
\newblock In \emph{ECCV}, 2024.

\bibitem[Umeyama(1991)]{umeyama1991least}
Shinji Umeyama.
\newblock Least-squares estimation of transformation parameters between two
  point patterns.
\newblock \emph{IEEE Transactions on Pattern Analysis and Machine
  Intelligence}, 13\penalty0 (4):\penalty0 376--380, 1991.

\bibitem[Wang et~al.(2025{\natexlab{a}})Wang, Xu, Ong, and
  Pollefeys]{casdiffmvs}
Fangjinhua Wang, Qingshan Xu, Yew-Soon Ong, and Marc Pollefeys.
\newblock Lightweight and accurate multi-view stereo with confidence-aware
  diffusion model.
\newblock \emph{IEEE Transactions on Pattern Analysis and Machine
  Intelligence}, 47\penalty0 (11):\penalty0 10060--10075, 2025{\natexlab{a}}.

\bibitem[Wang and Agapito(2025)]{wang2025spann3r}
Hengyi Wang and Lourdes Agapito.
\newblock 3d reconstruction with spatial memory.
\newblock In \emph{3DV}, 2025.

\bibitem[Wang et~al.(2023)Wang, Rupprecht, and Novotny]{wang2023posediffusion}
Jianyuan Wang, Christian Rupprecht, and David Novotny.
\newblock Posediffusion: Solving pose estimation via diffusion-aided bundle
  adjustment.
\newblock In \emph{ICCV}, 2023.

\bibitem[Wang et~al.(2024{\natexlab{a}})Wang, Karaev, Rupprecht, and
  Novotny]{wang2024vggsfm}
Jianyuan Wang, Nikita Karaev, Christian Rupprecht, and David Novotny.
\newblock {VGGSfM}: Visual geometry grounded deep structure from motion.
\newblock In \emph{CVPR}, 2024{\natexlab{a}}.

\bibitem[Wang et~al.(2025{\natexlab{b}})Wang, Chen, Karaev, Vedaldi, Rupprecht,
  and Novotny]{wang2025vggt}
Jianyuan Wang, Minghao Chen, Nikita Karaev, Andrea Vedaldi, Christian
  Rupprecht, and David Novotny.
\newblock {VGGT}: Visual geometry grounded transformer.
\newblock In \emph{CVPR}, 2025{\natexlab{b}}.

\bibitem[Wang et~al.(2026{\natexlab{a}})Wang, Chen, Zhang, Karaev,
  Sch{\"o}nberger, Labatut, Bojanowski, Novotny, Vedaldi, and
  Rupprecht]{wang2026vggtomega}
Jianyuan Wang, Minghao Chen, Shangzhan Zhang, Nikita Karaev, Johannes
  Sch{\"o}nberger, Patrick Labatut, Piotr Bojanowski, David Novotny, Andrea
  Vedaldi, and Christian Rupprecht.
\newblock {VGGT-$\Omega$}.
\newblock In \emph{CVPR}, 2026{\natexlab{a}}.
\newblock arXiv:2605.15195.

\bibitem[Wang et~al.(2025{\natexlab{c}})Wang, Xu, Dong, Deng, Xiang, Lv, Sun,
  Tong, and Yang]{moge2}
Ruicheng Wang, Sicheng Xu, Yue Dong, Yu Deng, Jianfeng Xiang, Zelong Lv,
  Guangzhong Sun, Xin Tong, and Jiaolong Yang.
\newblock {MoGe-2}: Accurate monocular geometry with metric scale and sharp
  details.
\newblock In \emph{NeurIPS}, 2025{\natexlab{c}}.

\bibitem[Wang et~al.(2024{\natexlab{b}})Wang, Leroy, Cabon, Chidlovskii, and
  Revaud]{wang2024dust3r}
Shuzhe Wang, Vincent Leroy, Yohann Cabon, Boris Chidlovskii, and J{\'e}r{\^o}me
  Revaud.
\newblock {DUSt3R}: Geometric 3d vision made easy.
\newblock In \emph{CVPR}, 2024{\natexlab{b}}.

\bibitem[Wang et~al.(2026{\natexlab{b}})Wang, Zhou, Zhu, Chang, Zhou, Li, Chen,
  Pang, Shen, and He]{wang2026pi3}
Yifan Wang, Jianjun Zhou, Haoyi Zhu, Wenzheng Chang, Yang Zhou, Zizun Li, Junyi
  Chen, Jiangmiao Pang, Chunhua Shen, and Tong He.
\newblock $\pi^3$: Permutation-equivariant visual geometry learning.
\newblock In \emph{ICLR}, 2026{\natexlab{b}}.

\bibitem[Wu et~al.(2023)Wu, Zhang, Fu, Wang, Ren, Pan, Wu, Yang, Wang, Qian,
  Lin, and Liu]{wu2023omniobject3d}
Tong Wu, Jiarui Zhang, Xiao Fu, Yuxin Wang, Jiawei Ren, Liang Pan, Wayne Wu,
  Lei Yang, Jiaqi Wang, Chen Qian, Dahua Lin, and Ziwei Liu.
\newblock {OmniObject3D}: Large-vocabulary {3D} object dataset for realistic
  perception, reconstruction and generation.
\newblock In \emph{CVPR}, 2023.

\bibitem[Wu et~al.(2024)Wu, Shi, Liu, Liao, Qiu, Yuan, Gu, Dong, Cui, and
  Han]{han2024mvimgnet2}
Yushuang Wu, Luyue Shi, Haolin Liu, Hongjie Liao, Lingteng Qiu, Weihao Yuan,
  Xiaodong Gu, Zilong Dong, Shuguang Cui, and Xiaoguang Han.
\newblock {MVImgNet2.0}: A larger-scale dataset of multi-view images.
\newblock \emph{ACM Transactions on Graphics}, 43\penalty0 (6):\penalty0
  173:1--173:16, 2024.

\bibitem[Xia et~al.(2024)Xia, Fu, Liu, and Wang]{xia2024wildrgbd}
Hongchi Xia, Yang Fu, Sifei Liu, and Xiaolong Wang.
\newblock Rgbd objects in the wild: Scaling real-world 3d object learning from
  rgb-d videos.
\newblock In \emph{CVPR}, 2024.

\bibitem[Xiang et~al.(2014)Xiang, Mottaghi, and Savarese]{xiang2014pascal3d}
Yu Xiang, Roozbeh Mottaghi, and Silvio Savarese.
\newblock Beyond pascal: A benchmark for 3d object detection in the wild.
\newblock In \emph{WACV}, 2014.

\bibitem[Xiang et~al.(2016)Xiang, Kim, Chen, Ji, Choy, Su, Mottaghi, Guibas,
  and Savarese]{xiang2016objectnet3d}
Yu Xiang, Wonhui Kim, Wei Chen, Jingwei Ji, Christopher Choy, Hao Su, Roozbeh
  Mottaghi, Leonidas Guibas, and Silvio Savarese.
\newblock {ObjectNet3D}: A large scale database for {3D} object recognition.
\newblock In \emph{ECCV}, 2016.

\bibitem[Xiangli et~al.(2025)Xiangli, Cai, Chen, Byrne, and
  Snavely]{xiangli2025doppelgangerspp}
Yuanbo Xiangli, Ruojin Cai, Hanyu Chen, Jeffrey Byrne, and Noah Snavely.
\newblock Doppelgangers++: Improved visual disambiguation with geometric 3d
  features.
\newblock In \emph{CVPR}, 2025.

\bibitem[Yang et~al.(2025)Yang, Sax, Liang, Henaff, Tang, Cao, Chai, Meier, and
  Feiszli]{yang2025fast3r}
Jianing Yang, Alexander Sax, Kevin~J. Liang, Mikael Henaff, Hao Tang, Ang Cao,
  Joyce Chai, Franziska Meier, and Matt Feiszli.
\newblock {Fast3R}: Towards {3D} reconstruction of 1000+ images in one forward
  pass.
\newblock In \emph{CVPR}, 2025.

\bibitem[Yao et~al.(2020)Yao, Luo, Li, Zhang, Ren, Zhou, Fang, and
  Quan]{yao2020blendedmvs}
Yao Yao, Zixin Luo, Shiwei Li, Jingyang Zhang, Yufan Ren, Lei Zhou, Tian Fang,
  and Long Quan.
\newblock {BlendedMVS}: A large-scale dataset for generalized multi-view stereo
  networks.
\newblock In \emph{CVPR}, 2020.

\bibitem[Yeshwanth et~al.(2023)Yeshwanth, Liu, Nie{\ss}ner, and
  Dai]{yeshwanth2023scannetpp}
Chandan Yeshwanth, Yueh-Cheng Liu, Matthias Nie{\ss}ner, and Angela Dai.
\newblock Scannet++: A high-fidelity dataset of 3d indoor scenes.
\newblock In \emph{ICCV}, 2023.

\bibitem[Yu et~al.(2023)Yu, Xu, Zhang, Liu, Ye, Wu, Yan, Zhu, Xiong, Liang,
  Chen, Cui, and Han]{yu2023mvimgnet}
Xianggang Yu, Mutian Xu, Yidan Zhang, Haolin Liu, Chongjie Ye, Yushuang Wu,
  Zizheng Yan, Chenming Zhu, Zhangyang Xiong, Tianyou Liang, Guanying Chen,
  Shuguang Cui, and Xiaoguang Han.
\newblock {MVImgNet}: A large-scale dataset of multi-view images.
\newblock In \emph{CVPR}, 2023.

\bibitem[Yu et~al.(2026)Yu, Gao, Zhang, Qiu, Zhao, Peng, Yan, Qiu, Zhu, Dong,
  Cao, and Shen]{vigeo}
Zhu Yu, Jingnan Gao, Runmin Zhang, Lingteng Qiu, Zhengyi Zhao, Rui Peng, Yichao
  Yan, Kejie Qiu, Siyu Zhu, Zilong Dong, Si-Yuan Cao, and Hui-Liang Shen.
\newblock Towards consistent video geometry estimation.
\newblock \emph{arXiv preprint arXiv:2605.30060}, 2026.

\bibitem[Zhang et~al.(2025)Zhang, Wang, Xu, Xue, Rupprecht, Zhou, Shen, and
  Wetzstein]{zhang2025flare}
Shangzhan Zhang, Jianyuan Wang, Yinghao Xu, Nan Xue, Christian Rupprecht,
  Xiaowei Zhou, Yujun Shen, and Gordon Wetzstein.
\newblock {FLARE}: Feed-forward geometry, appearance and camera estimation from
  uncalibrated sparse views.
\newblock In \emph{CVPR}, 2025.

\bibitem[Zhang et~al.(2026)Zhang, Tung, Cai, Fouhey, and
  Averbuch-Elor]{zhang2025megaunscene}
Yiwen Zhang, Joseph Tung, Ruojin Cai, David Fouhey, and Hadar Averbuch-Elor.
\newblock Emergent extreme-view geometry in {3D} foundation models.
\newblock In \emph{CVPR}, 2026.

\end{thebibliography}
